\documentclass[review]{elsarticle}

\let\today\relax
\makeatletter
\def\ps@pprintTitle{%
	\let\@oddhead\@empty
	\let\@evenhead\@empty
	\def\@oddfoot{\footnotesize\itshape
		{Preprint} \hfill\today}%
	\let\@evenfoot\@oddfoot
}
\makeatother

\usepackage{hyperref}
\usepackage{amsmath,amssymb,amsfonts}
\usepackage{algorithmic}
\usepackage{graphicx}
\usepackage{textcomp}

\usepackage{makecell,multirow,subfig}
\renewcommand{\tablename}{Table}
\renewcommand{\figurename}{Figure}
\newcommand{\figurecolumn}{3.40in}
\newcommand{\figureraster}{0.25}
\newcommand{\figureconfusion}{0.20}

\renewcommand{\arraystretch}{1.2}                

\newcommand{\indicator}[1]{{\mathbf{1}}\left( #1 \right)}

\newcommand{\bydef}{\triangleq}

\DeclareMathOperator*{\argmin}{arg\,min}

\journal{}
\begin{document}

\begin{frontmatter}
	\title{
		Iteration over event space in time-to-first-spike spiking neural networks
		for Twitter bot classification%
		\tnoteref{t1}
	}
	\tnotetext[t1]{This work was supported by the Polish National Center of Science under Grant DEC-2017/27/B/ST7/03082.}
	
	\author[1]{Mateusz Pabian\corref{cor1}}
	\ead{pabian@agh.edu.pl}
	
	\author[1]{Dominik Rzepka}
	
	\author[2,1,3]{Miros\l{}aw Pawlak}
	
	\cortext[cor1]{Corresponding author}
	
	\affiliation[1]{
		organization={Department of Measurement and Electronics, AGH University of Krakow},
		addressline={al. Mickiewicza 30},
		postcode={30-059},
		city={Kraków},
		country={Poland}
	}
	
	\affiliation[2]{
		organization={Department of Electrical and Computer Engineering, University of Manitoba},
		addressline={75 Chancellors Circle},
		postcode={R3T 5V6},
		city={Winnipeg},
		country={Canada}
	}
	
	\affiliation[3]{
		organization={Information Technology Institute, University of Social Sciences},
		addressline={ul. Sienkiewicza 9},
		postcode={90-113},
		city={Łódź},
		country={Poland}
	}
		
	\begin{abstract}
		This study proposes a framework that extends existing time-coding time-to-first-spike
spiking neural network~(SNN) models to allow processing information changing over time.
We explain spike propagation through a model with multiple input and output spikes at
each neuron, as
well as design training rules for end-to-end backpropagation. This strategy enables us to 
process information changing over time. The model is trained and evaluated on a Twitter bot 
detection task where the time of events (tweets and retweets) is the primary carrier of 
information. This task was chosen to evaluate how the proposed~SNN deals with spike train 
data composed of hundreds of events occurring at timescales differing by almost five 
orders of magnitude. The impact of various parameters on model properties, performance and 
training-time stability is analyzed.

	\end{abstract}
	
\end{frontmatter}

\section{Introduction}
\label{sec:01_introduction}

Spiking neural networks~(SNN) differ from classic artificial neural networks in how the
signal is represented and propagated through the network. This in turn determines their 
applicability to event-driven types of data (i.e., characterized by event occurrence as 
the primary information carrier, or data that can be represented in this form by a proper 
encoding). SNN process data using impulses which 
are asynchronously propagated through the entire
network~\cite{Pfeiffer2018, eshraghian2023training}. 
This processing scheme stands in contrast to 
the classic artificial neural networks which require that all neurons within a single 
layer must finish their computation before the signal flows to the subsequent layer. 
Overall, the aim of SNN is to model biological networks in a much more principled way. 
The training rules for the~SNN generally fall into one of three categories: network 
conversion (mapping each component of the source network to its spiking 
equivalent)~\cite{Rueckauer2017, midya2019artificial, stockl2021optimized},
synaptic-plasticity-aware training (taking advantage of long-term potentiation and 
long-term depression effects)~\cite{falez2019unsupervised, mozafari2019bio}, 
or training with backpropagation~\cite{
    Mostafa2018, wu2018spatio, rasmussen2019nengodl%
}. The latter
approach necessitates formulating custom training rules that take into account the fact 
that the activation function of biological neurons -- ``all-or-nothing'' principle -- is 
not differentiable. Hybrid conversion-backpropagation approaches have also been 
explored~\cite{guo2023joint}.

There are several factors that inhibit research on the~SNN. One of them is the 
availability of event data. In case such dataset is unavailable (due to its proprietary
nature or when the original signal exists in the analog domain), researchers can opt to
transform existing sets of data to an event structure~\cite{
	Garrick2015, doutsi2021dynamic%
}. Regardless of how the event data is obtained, it is possible to compare 
the~SNN with nonspiking models trained on the same task. Another inhibiting 
factor is the computational complexity of simulating the~SNN on typical hardware. 
Fortunately, the research on spiking neural networks goes hand-in-hand with the 
development of dedicated neuromorphic hardware~\cite{javanshir2022advancements}.

In this paper we use backpropagation to train an SNN based on the time-to-first-spike 
neurons introduced in~\cite{Mostafa2018}. We address one of the limitations of this model 
related to the infinitely long neuron refractory period $\tau_{\text{ref}}$. This 
shortcoming means that every neuron in the model can elicit at most one spike during a 
single input example presentation. Overcoming this limitation allows us to design a model 
that is able to process information changing over time. Crucially, the proposed signal 
propagation and network training rules are truly event-centric, evaluating the state of 
the model at each event rather than at fixed points in time determined by the simulation 
grid. We evaluate our approach by 
training a classification network on real-life dataset of legitimate and automated 
Twitter user activity~\cite{mazza2019rtbust}. We choose this data for several reasons. 
First of all, each record is described only in terms of the time of event without any 
auxiliary information (such as tweet content and sentiment). Secondly, spike trains 
formed for each record are several hundred events long. Lastly, the events occur at 
timescales differing by many orders of magnitude. This allows us to assess the 
feasibility of the proposed approach in scenarios commonly encountered when processing 
this type of data.

This paper is structured as follows. In 
Sections~\ref{subsec:01_introduction/backprop} and~\ref{subsec:01_introduction/twitter} 
we give a brief overview of backpropagation-based 
algorithms for training time-coding SNN, as well as introduce the Twitter bot 
detection problem. Section~\ref{sec:02_methods} 
describes the signal propagation and training rules for the proposed SNN, with the main 
contributions outlined in 
Sections~\ref{subsec:02_methods/mimo} and~\ref{subsec:02_methods/loss}. The dataset
properties, training data selection and preprocessing steps are described in
Section~\ref{sec:03_experiment}. Finally, we summarize our experimental results in 
Section~\ref{sec:04_results}, showing the impact of model parameters on network properties 
and performance.

\subsection{Related works on backpropagation-based training of SNN}
\label{subsec:01_introduction/backprop}

Training artificial spiking neural networks with backpropagation leverages existing 
methods, algorithms and best practices developed for  artificial nonspiking neural 
networks and deep learning. In doing so it forgoes biological 
plausibility, evident in training methods based on spike-timing-dependent plasticity 
effects~\cite{falez2019unsupervised}.

One of the first proposed 
gradient-descent-based learning rules for timing-encoding networks was the SpikeProp 
algorithm~\cite{bohte2002error} and its variants. It established key 
traits of backpropagation algorithms present in subsequent works. In particular, the 
algorithm has the following characteristics: 1)~the SNN is simulated 
over a finite time window with a fixed time step, recording changes in the internal state 
(membrane voltage, synaptic current) of all neurons during the forward pass and 
2)~approximating (smoothing) the spike-generation function to avoid a discontinuous 
derivative (unless this discontinuity is ignored and treated as noise, as
in~\cite{Lee2016}). Backpropagation-based approaches for training time-coding SNN can
thus be divided into two categories, depending on how much information from the forward 
pass is needed to compute the backward pass. In event-driven learning, the error is 
propagated only through 
spikes~\cite{huh2018gradient, zhang2020temporal, perez2021sparse, zhu2022training}. Most
notably, EventProp~\cite{wunderlich2021event} defines exact gradients and can be applied 
to neuron models without an analytical expression for the postsynaptic potential kernels. 
Nevertheless, similarly to other existing algorithms, EventProp still requires simulating 
the network in the forward pass in order to compute postsynaptic 
events. As these algorithms propagate the error only through spike 
events, they are prone to fail to converge when the network does not generate enough 
events to process the signal end-to-end.

The other category that stands in contrast to event-driven learning is RNN-like learning 
(named after its similarity to nonspiking artificial Recurrent Neural Networks), 
in which the error information is also propagated through computation time steps which 
did not elicit a spike~\cite{wu2018spatio, zheng2021going}. The 
SuperSpike~\cite{zenke2018superspike} and 
SLAYER~\cite{shrestha2018slayer} models are examples of such algorithms. 
Surrogate gradient methods, commonly used in RNN-like learning, are an alternative 
approach for overcoming the discontinuous derivative of the spike-generating 
function~\cite{neftci2019surrogate}. Typically, a standard backpropagation-through-time 
algorithm~\cite{werbos1990backpropagation} is used, as in RNN, with one minor 
modification: a continuously differentiable 
function is used in the backward pass as a surrogate of the spike-generation function 
derivative. Finding the optimal surrogate gradient function is a topic of an ongoing 
research~\cite{yin2021accurate}. Despite achieving a remarkable 
success in training deep SNN~\cite{fang2021deep, kim2022beyond}, surrogate gradient 
methods represent an even further departure from energy-efficient, biologically-inspired 
learning.

Recently, there has been some research conducted on single-spike time-to-first-spike SNN 
that specify learning rules which do not require simulating the network over 
time~\cite{Mostafa2018, kheradpisheh2020temporal, zhou2021temporal}.
Concretely, \cite{Mostafa2018}~trains an~SNN with 
simplistic Integrate-and-Fire (IF) neurons by deriving locally exact gradients of the 
spike-generating function. A similar idea is explored in~\cite{kheradpisheh2020temporal}
where the instantaneous synaptic current kernel function is used instead of an
exponentially decaying kernel. Furthermore, 
\cite{zhou2021temporal}~applies popular deep learning techniques such as max-pooling and 
batch normalization to this model type, which alleviates issues with training deeper 
neural architectures. However, to the best of our knowledge, there has been no research 
that shows how this model can be applied to actual spike trains rather than single spikes.

In this work we derive an algorithm that extends the aforementioned single-spike 
time-to-first-spike SNN training rules to neurons observing and generating multiple
spike trains. In doing so we not only infer locally exact gradients of the 
spike-generating function, like the EventProp algorithm does, but also express the entire 
computation in terms of iteratively computing successive spikes (first, second, third, 
etc.). The latter property stands in stark contrast to other works which simulate the 
state of the entire network over a finite time window with a fixed time step.

\subsection{An overview of Twitter bot detection}
\label{subsec:01_introduction/twitter}

Social bots are automated agents that interact with humans and mimic human 
behavior~\cite{ferrara2016rise}. In social media environment, such bots may aggregate 
contents from various sources, automatically respond to custom queries, or even generate content 
that satisfies some constraints. However, some bot behavior and intent is purposefully 
misleading or downright malicious~\cite{cresci2017paradigm}.

Bot detection systems can be divided into three categories: crowdsourcing, feature-based 
classification, and social network analysis. Crowdsourcing techniques rely on the ability of 
hired human experts and volunteers to manually annotate bots based on their profiles, or on 
sending survey requests to the users and analyzing their 
replies~\cite{cresci2017paradigm}. Feature-based classification aims to automate the process by 
analyzing user-level features, usually by aggregating data pertaining to the user itself and 
to their activity~\cite{davis2016botornot, rodriguez2020one}. 
It is also possible 
to perform unsupervised clustering of users based on the temporal similarity of their 
activity, regardless of their relationship in the social network~\cite{mazza2019rtbust}. 
This circumvents the time-consuming and costly process of labeling the data. Additionally, 
the unsupervised systems do not become outdated when new generations of bots, exhibiting 
patterns of behavior not present during model training, become more prevalent. However, they 
treat all unclustered examples as legitimate users, limiting their ability to detect bots 
exhibiting irregular behavior. The last of the three bot 
detection categories expands the scope of analysis to operate on a large group of users at 
once~\cite{minnich2017botwalk, balaanand2019enhanced}. Similarly to the 
unsupervised user-level methods, they do not become obsolete due to data 
drift~\cite{minnich2017botwalk}. However, given that they analyze the actual community 
structure formed by users, they take more time to process information, making them more 
difficult to operate at scale~\cite{cao2012aiding}.

In this work we focus on analyzing user data in terms of tweet and retweet actions as the 
only information available to the model. A tweet is the original message that can be 
re-broadcasted (i.e., retweeted) by other users. While it 
is more common to analyze retweets with  group-based 
methods~\cite{chavoshi2016identifying, gupta2019malreg}, automated 
behavior can also be made evident by analyzing user-level patterns of retweet 
activity~\cite{pan2016discriminating, dutta2018retweet, mazza2019rtbust}. The latter is the
approach followed in our study.

\section{Methods \& algorithms}
\label{sec:02_methods}

\subsection{Signal propagation in time-to-first-spike SNN}
\label{subsec:02_methods/backprop}


This study focuses on a specific type of SNN that is sensitive to the timing of input 
events and not their rate. Therefore, let us briefly summarize the model first 
described by Mostafa~\cite{Mostafa2018}. This type of network uses IF~neurons with 
exponentially decaying synaptic current kernels. The membrane 
voltage of a IF neuron with~$C$ presynaptic neuron connections is governed by the 
differential equation
\begin{equation}
	\frac{dV(t)}{dt} = \sum_{c=1}^{C} w_c i_c(t)
	\label{eqn:02_methods/backprop/if_dynamics} \,,
\end{equation}
where~$w_c$ is the weight associated with the~$c$-th synapse (channel), and~$i_c(t)$ 
is the synaptic current driving signal. Assuming that every presynaptic neuron 
observes a single event at time~$t_c$, the presynaptic current is
\begin{equation}
	i_c(t) = \exp \left( -\frac{t-t_c}{\tau_{\text{syn}}} \right) u(t-t_c)
	\label{eqn:02_methods/backprop/if_current} \,,
\end{equation}
with~$\tau_{\text{syn}}$ being the synaptic current time constant and~$u(t)$ being the 
step function. The solution to the system of equations defined 
by~\eqref{eqn:02_methods/backprop/if_dynamics}
and~\eqref{eqn:02_methods/backprop/if_current} is given by
\begin{equation}
	V(t) = 
	V_0 + \tau_{\text{syn}} 
	\sum_{c=1}^{C} w_c \left[ 1 - \exp \left( 
	-\frac{t-t_c}{\tau_{\text{syn}}} 
	\right) \right] u(t-t_c)
	\label{eqn:02_methods/backprop/if_solution} \,,
\end{equation}
where~$V_0=V(0)$ is the initial membrane voltage. The neuron is said to fire at 
time~$t_{\text{out}}$ if the voltage exceeds a threshold~$V_{\text{thr}}$,  
i.e.,~\mbox{
	$t_{\text{out}} = \argmin_t V(t) \geq V_{thr}$
}, after which the IF neuron becomes unresponsive to input signals for some time, 
called the refractory period~$\tau_{\text{ref}}$. Once the refractory period subsides, 
the voltage is reset to zero and the 
summation over impulses in~\eqref{eqn:02_methods/backprop/if_solution} can resume.
For now let us consider the case~\mbox{
	$\tau_{\text{ref}} \to \infty$
}, meaning that the neuron is unable to respond to new inputs following output spike 
generation. This is an implicit assumption in single-spike 
SNN~\cite{Mostafa2018, kheradpisheh2020temporal, zhou2021temporal}.

Assuming without the loss of generality~$V_0=0$ and that there exists a subset of 
input spikes that cause the postsynaptic neuron to fire
\begin{equation}
	Q = \lbrace c: t_c < t_{\text{out}} \rbrace
	\label{eqn:02_methods/backprop/causal_set} \,,
\end{equation}
i.e. the \emph{causal set} of input spikes, the solution for~$t_{\text{out}}$ is given 
in the implicit form
\begin{equation}
	z_{\text{out}} = 
	\frac{\sum_{c \in Q} w_c z_c}{\sum_{c \in Q} w_c - 
		\frac{V_{\text{thr}}}{\tau_{\text{syn}}}}
	\label{eqn:02_methods/backprop/if_zout} \,,
\end{equation}
where
\begin{equation}
	z_c(t) = \exp \left( \frac{t_c}{\tau_{\text{syn}}} \right),
	\enspace
	z_\text{out}(t) = \exp \left( \frac{t_\text{out}}{\tau_{\text{syn}}} \right)
	\label{eqn:02_methods/backprop/z_transform} \,.
\end{equation}
For completeness we assign~\mbox{
	$z_{\text{out}} = \infty$ when $Q=\varnothing$
}. A necessary condition for the postsynaptic neuron to fire is that the sum of 
weights of the causal set of neurons be strictly larger than the scaled threshold 
voltage~\cite{Mostafa2018}, i.e., that
\begin{equation}
	1 \leq z_{\text{out}} < \infty
	\iff
	\sum_{c \in Q}w_c > \frac{V_{\text{thr}}}{\tau_{\text{syn}}}
	\label{eqn:02_methods/backprop/necessary_condition} \,.
\end{equation}
Note that the condition in~\eqref{eqn:02_methods/backprop/necessary_condition} assures 
that the right-hand side of~\eqref{eqn:02_methods/backprop/if_zout} is positive.
The formula~\eqref{eqn:02_methods/backprop/if_zout} is differentiable with
respect to transformed input spike times~$\lbrace z_c \rbrace$ and synaptic 
weights~$\lbrace w_c \rbrace$ with partial derivatives
\begin{equation}
	\frac{\partial z_{\text{out}}}{\partial z_c}=
	\begin{cases}
		\frac{w_c}{\sum_{c \in Q}w_c - \frac{V_{\text{thr}}}{\tau_{\text{syn}}}} 
		& \text{if $c \in Q$} \\
		0 
		& \text{otherwise}
	\end{cases}
	\label{eqn:02_methods/backprop/deriv_zc} \,,
\end{equation}
\begin{equation}
	\frac{\partial z_{\text{out}}}{\partial w_c}=
	\begin{cases}
		\frac{z_c-z_{\text{out}}}{\sum_{c \in Q}w_c - 	
			\frac{V_{\text{thr}}}{\tau_{\text{syn}}}} 
		& \text{if $c \in Q$} \\
		0 
		& \text{otherwise}
	\end{cases}
	\label{eqn:02_methods/backprop/deriv_wc} \,,
\end{equation}
therefore it can be used to train a spiking network using the backpropagation 
algorithm.

\subsection{Multiple-input, multiple-output (MIMO) SNN}
\label{subsec:02_methods/mimo}

\begin{figure}
	\includegraphics[width=\figurecolumn]{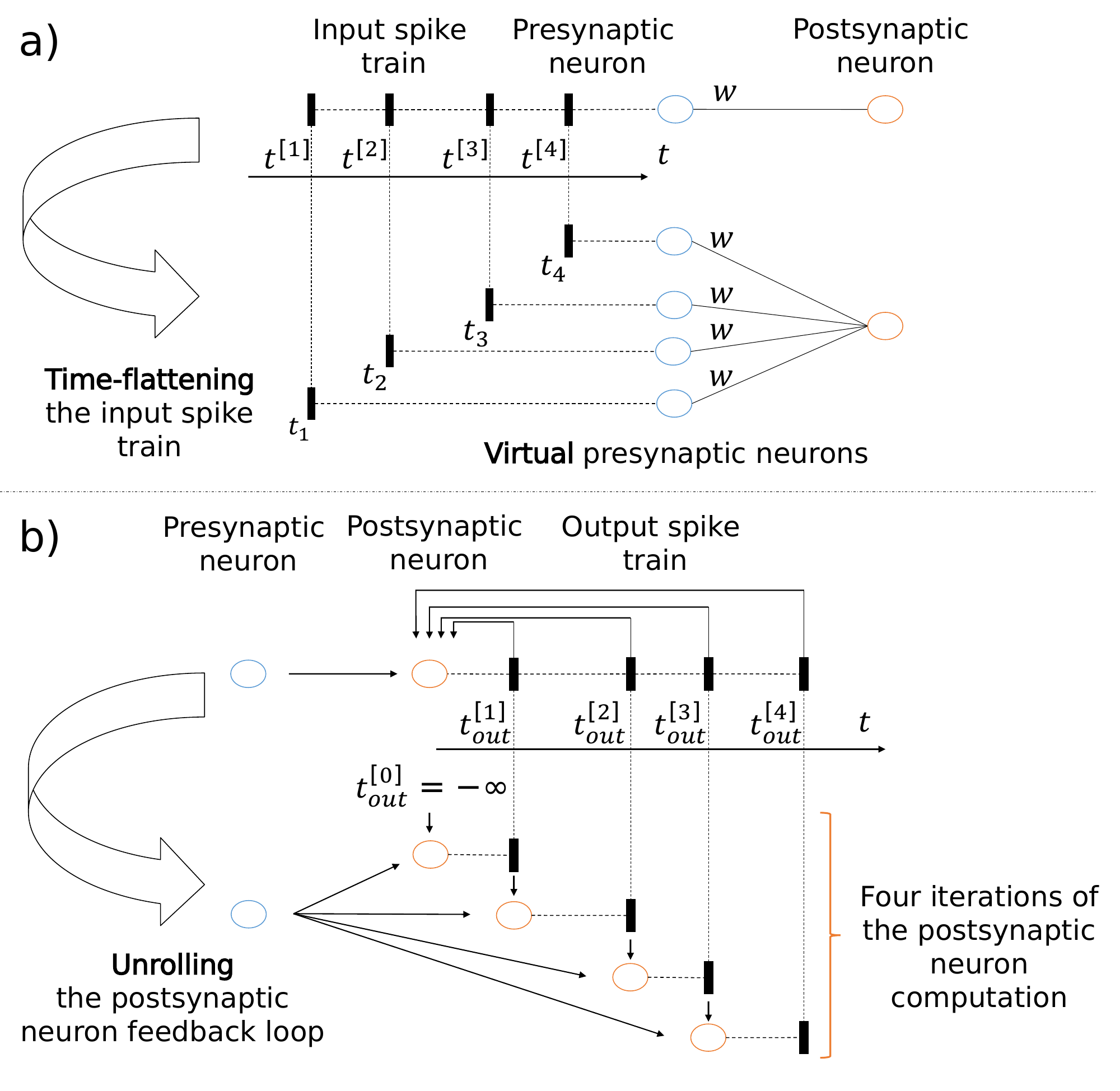}
	\centering
	\caption{
		Signal propagation rules in MIMO~SNN. The flow of time is represented by an axis 
		going from left to right, i.e., the earliest spike is at the left side of each 
		subfigure. 
		a)~A presynaptic neuron that observes multiple input spikes can be represented as 
		multiple virtual presynaptic neurons, each observing a single spike. All virtual 
		presynaptic neurons have the same weight between them and the postsynaptic neuron, 
		identical to the weight associated with the original connection before 
		time-flattening.
		b)~The ability of the postsynaptic neuron to produce spikes depends on all input 
		spike trains from presynaptic neurons, as well as the previously generated 
		output. This feedback loop imposed by the spike causality principle can be 
		unrolled over time, where the postsynaptic neuron 
		computation is repeated with the same input spike trains, but for different 
		timestamps of the previously generated event. This cascade proceeds until it is 
		impossible for the postsynaptic neuron to generate an output spike. The implicit 
		output spike at $t_{\text{out}}^{[0]}=-\infty$ designates the initial state of 
		the postsynaptic neuron, i.e., it has not generated a spike yet.}
	\label{fig:02_methods/mimo/design}
\end{figure}

The model summarized in the previous Section is a time-coding single-spike SNN, 
unable to process information
changing over time. It assumes that every neuron in the model, including input 
neurons, can elicit at most one spike (by implicitly setting 
$\tau_{\text{ref}}=\infty$). Our goal is to extend the signal propagation rules of
the aforementioned model so that its hidden and output neurons are also able to
produce multiple spikes each. For brevity, we use a \emph{MIMO network} descriptor 
throughout this Section to refer to the model that supports signal propagation with 
multiple inputs and multiple outputs. In case of multiple events arriving at the input 
synapse, the equation~\eqref{eqn:02_methods/backprop/if_dynamics} becomes
\begin{equation}
	\frac{dV(t)}{dt} 
	= \sum_{c=1}^{C} w_c \sum_{j=1}^{T_c} i_c^{[j]}(t) 
	= \sum_{c=1}^{C} \sum_{j=1}^{T_c} w_c^{[j]} i_c^{[j]}(t)
	\label{eqn:02_methods/mimo/if_dynamics_multi}\,,
\end{equation}
where~$T_c$ is the number of events observed in channel~$c$. Note that this formula
explicitly highlights the fact that every event~$t_c^{[j]}$ of channel~$c$ is
associated with the same weight \mbox{$\lbrace j: w_c^{[j]} = w_c \rbrace$}.
Therefore, a single channel with~$T_c$ events is equivalent 
to~$T_c$ \emph{virtual} (or \emph{time-flattened}) channels with a single event 
each. Introducing a new index \mbox{$\lbrace k: k = 1, \ldots, K \rbrace$} over 
these virtual input channels such that \mbox{$K = \sum_{c=1}^{C}T_c$} we obtain the 
following representation of~\eqref{eqn:02_methods/mimo/if_dynamics_multi}
\begin{equation}
	\frac{dV(t)}{dt} 
	= \sum_{k=1}^{K} w_{k} i_{k}(t)
	\label{eqn:02_methods/mimo/if_dynamics_multi_repr}
\end{equation}
which is identical to~\eqref{eqn:02_methods/backprop/if_dynamics}.
It follows that the equations for forward and backward signal propagation through the
network introduced 
in~\eqref{eqn:02_methods/backprop/if_zout}-\eqref{eqn:02_methods/backprop/deriv_wc}
still hold for the layer observing
inputs spiking over time, provided that the indices are substituted where
appropriate. The proposed idea of time-flattening the input signal and projecting
input layer weights is presented in
\figurename~\ref{fig:02_methods/mimo/design}a.

\begin{figure}
	\subfloat[$\tau_{\text{ref}}=0.1$]{\includegraphics[scale=\figureraster]
		{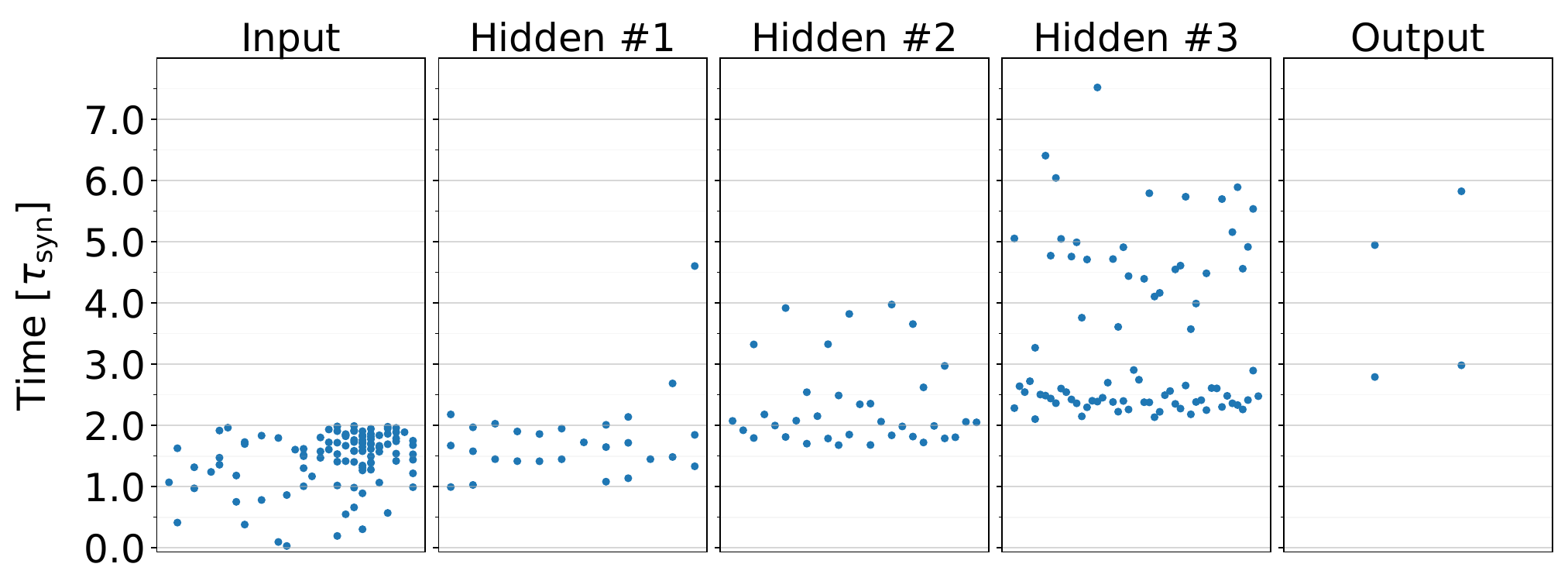}
	}\quad\quad
	\subfloat[$\tau_{\text{ref}}=1$]{\includegraphics[scale=\figureraster]
		{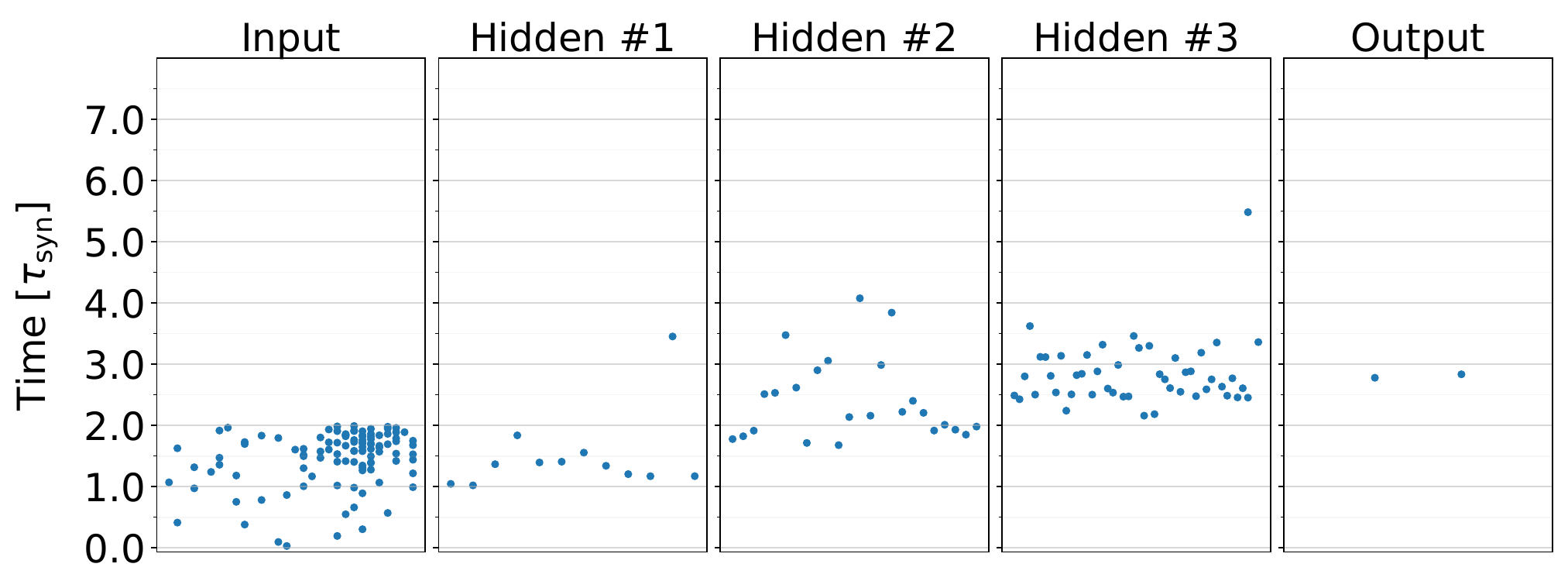}
	}
	\centering
	\caption{
		Spike raster plot for models trained with two different values 
		of~$\tau_{\text{ref}}$ responding to the same input example.
	}
	\label{fig:02_methods/mimo/raster}
\end{figure}

Furthermore, allowing each neuron to spike more than once requires setting a finite
nonnegative refractory period~$\tau_{\text{ref}}$. In that case the causal set of
input spikes~\eqref{eqn:02_methods/backprop/causal_set} becomes
\begin{equation}
	Q_m = 
	\begin{cases}
		\lbrace k: t_{k} < t_{\text{out}}^{[m]} \rbrace & \text{if $m=1$} \\
		\lbrace k: t_{\text{out}}^{[m-1]} + \tau_{\text{ref}} < t_{k} < 
		t_{\text{out}}^{[m]} \rbrace & \text{otherwise}
	\end{cases}
	\label{eqn:02_methods/mimo/causal_set_modified} \,,
\end{equation}
where \mbox{$\lbrace m: m = 1, \ldots, M \rbrace$} is the index over the sequence of 
neuron output spikes. Computing the~$m$-th output spike time can be done 
iteratively, noting that~$M$ is at most equal to~$K$ (the neuron cannot spike more 
often than once per input spike). The iterative computation over~$m$ can be stopped 
early when an empty causal set is encountered (\mbox{$Q_m=\varnothing$}). Taking 
this new definition of a causal set into consideration, the implicit formula 
for~$t_{\text{out}}^{[m]}$ becomes
\begin{equation}
	z_{\text{out}}^{[m]} 
	= \frac{\sum_{k \in Q_m} w_{k} z_{k}}{\sum_{k \in Q_m} w_{k} - 
		\frac{V_{\text{thr}}}{\tau_{\text{syn}}}}
	\label{eqn:02_methods/mimo/if_zout_modified} \,,
\end{equation}
whereas the partial derivatives are
\begin{equation}
	\frac{\partial z_{\text{out}}^{[m]}}{\partial z_{k}}=
	\begin{cases}
		\frac{w_{k}}{\sum_{k \in Q_m} w_{k} - \frac{V_{\text{thr}}}{\tau_{\text{syn}}}} 
			& \text{if $k \in Q_m$} \\
		0 
			& \text{otherwise}
	\end{cases}
	\label{eqn:02_methods/mimo/deriv_zc_modified} \,,
\end{equation}
\begin{equation}
	\frac{\partial z_{\text{out}}^{[m]}}{\partial w_{k}}=
	\begin{cases}
		\frac{z_{k}-z_{\text{out}}^{[m]}}{\sum_{k \in Q_m} w_{k} - 
			\frac{V_{\text{thr}}}{\tau_{\text{syn}}}} 
			& \text{if $k \in Q_m$} \\
		0 
			& \text{otherwise}
	\end{cases}
	\label{eqn:02_methods/mimo/deriv_wc_modified} \,,
\end{equation}
and additionally
\begin{equation}
	\frac{\partial z_{\text{out}}^{[m]}}{\partial w_c}
	= \sum_{\lbrace k:\; w_k \bydef w_c\rbrace}\frac{\partial z_{\text{out}}^{[m]}}{\partial w_{k}}
	\label{eqn:02_methods/mimo/deriv_wc_modified_sum} \,,
\end{equation}
where the set $\lbrace k: w_k \bydef w_c\rbrace$ denotes the subset of all $k$ for 
which the virtual (time-flattened) weight $w_k$ corresponds to the original $w_c$.
Note the absence of any explicit dependence between two consecutive output spikes 
\mbox{$\lbrace t_{\text{out}}^{[m]}, t_{\text{out}}^{[m+1]} \rbrace$} in 
equations~
\eqref{eqn:02_methods/mimo/if_zout_modified}-\eqref{eqn:02_methods/mimo/deriv_wc_modified_sum},
or equivalently \mbox{$
	\forall_{m} \,
	\partial z_{\text{out}}^{[m+1]} / \partial z_{\text{out}}^{[m]}
	=0
$}. This is fundamentally different from signal propagation in the
backpropagation-through-time algorithm~\cite{werbos1990backpropagation}, and
is a direct result of the IF neuron's independence on its own history (i.e., it is 
memoryless). Instead, the gradients are simply summed over output spikes. This 
concept of iterating over the~$M$ output spikes is summarized in 
\figurename~\ref{fig:02_methods/mimo/design}b. Signal propagation through the MIMO~SNN 
model can be visualized using a spike raster plot, such as the in 
\figurename~\ref{fig:02_methods/mimo/raster}.

Overall, the proposed algorithm shows how to simulate and train a time-coding MIMO SNN  by 
time-flattening the presynaptic spike trains and unrolling the postsynaptic neuron
feedback loop imposed by the spike causality principle. This procedure is sufficient to
encode and process information with multiple spikes. However, it must be stressed that
the concepts described above and illustrated in
\figurename~\ref{fig:02_methods/mimo/design} specifically refer to operating the model on
conventional hardware. Once trained, the spiking network
(i.e, synaptic weights and neuron-specific hyperparameters) can be realized on existing 
neuromorphic hardware, in which case the virtual presynaptic neurons are no longer needed 
(hence the name) as reusing neurons for multiple input and output events is implied.

\subsection{MIMO SNN loss function}
\label{subsec:02_methods/loss}


The original model proposed by~\cite{Mostafa2018} was trained using a loss function 
with two components. The first one is a modification of the cross-entropy loss with 
softmax normalization
\begin{equation}
	L(z,y) = 
	- \sum_{p=1}^{P}y_p \ln \left(
	\frac{\exp\left(-z_p\right)}{\sum_{p=1}^{P} \exp\left(-z_p\right)}
	\right)
	\label{eqn:02_methods/loss/cross_entropy} \,,
\end{equation}
where:
\begin{itemize}
	\item $P$ - the number of output channels,
	\item $y_p$ - a binary indicator (0 or 1) of the desired output channel~$p$ 
	spiking first,
	\item $z_p$ - the transformed spike time of the $p$-th output channel.
\end{itemize}
This loss function encourages the model to use rank-order coding~\cite{thorpe1998rank} to 
represent the output, without explicitly specifying the time of each spike. The second 
component of the loss function is a spike regularization term, which promotes network 
spiking activity by ensuring a nonnegative denominator 
of~\eqref{eqn:02_methods/backprop/if_zout}
\begin{equation}
	R_{\text{spiking}} = \sum_{h=1}^{H} R_h
	\label{eqn:02_methods/loss/spike_penalty} \,,
\end{equation}
where \mbox{$
	R_h	= \max \left(
			0, V_{\text{thr}} / \tau_{\text{syn}} - \sum_{c=1}^{C_h} w_{ch} 
		\right)
$} and the index~$h$ runs over all neurons in the network~$H$. The overall loss 
function minimized by the model is
\begin{equation}
	L_{\text{total}}(z,y) = 
	\frac{1}{N} \sum_{n=1}^{N} L_n(z,y) + \gamma R_{\text{spiking}} 
	\label{eqn:02_methods/loss/loss} \,,
\end{equation}
where $L_n(z,y)$ is the cross-entropy loss~\eqref{eqn:02_methods/loss/cross_entropy}
for the $n$-th example of the batch of size~$N$, and $\gamma$~is a hyperparameter.


We posit that the penalty term should be applied only when ``the task is not solved''. For 
classification models this entails that the output of the network is different from the 
expected ground truth. Such approach would result in models that are more efficient in 
terms of the number of spikes needed to propagate the signal through the network, compared 
to the original approach (which makes all neurons in the network produce at least one 
spike). The proposed heuristic is thus
\begin{equation}
	R_{\text{spiking}}^{*} =
	\frac{1}{\sum_{n=1}^{N} \indicator{y_n \neq \hat{y}_n}}
	\sum_{n=1}^{N} \indicator{y_n \neq \hat{y}_n} R_{\text{spiking}}^{[n]}
	\label{eqn:02_methods/loss/dynamic_penalty} \,,
\end{equation}
where~$n$ runs over all examples withing a training minibatch~$N$;
$R_{\text{spiking}}^{[n]}$~is the spike-firing penalty 
term in~\eqref{eqn:02_methods/loss/spike_penalty},
and~$\mathbf{1}(A)$ is the indicator function. The argument of the indicator
function is a short-hand notation for the output of the network~$\hat{y}_n$ being
different from the expected ground truth~$y_n$. The proposed
heuristic~\eqref{eqn:02_methods/loss/dynamic_penalty} dynamically scales the
spike-firing penalty term during training - it acts as a mean penalty over the
minibatch examples if the network returns an incorrect output for all examples, and
it assigns an increasingly larger weight to each incorrectly predicted example as
the training progresses. For completeness
\begin{equation*}
	R_{\text{spiking}}^{*} =
	0 \quad \textrm{if}
	\quad \mathop{\forall}_{n \in \lbrace 1, \ldots, N \rbrace}
	\indicator{y_n \neq \hat{y}_n} = 0 \,.
\end{equation*}


Furthermore, some adjustments to the loss function components are necessary in order 
to use them in a MIMO~SNN context, particularly to the spike regularization term.
The formula~\eqref{eqn:02_methods/loss/spike_penalty} can be trivially extended to 
networks with inputs spiking over time by substituting the index over the input 
neurons~$c$ with an index over virtual input channels~$k$. We can also apply 
regularization to each neuron output spike~$m$ separately. Thus, the regularization 
term that considers the MIMO~SNN signal propagation rules is
\begin{equation}
	R_{\text{spiking}} = \sum_{h=1}^{H} \sum_{m=1}^{M_h} R_h
	\label{eqn:02_methods/loss/spike_penalty_virtual} \,,
\end{equation}
where $R_h$ is redefined as \mbox{$
	R_h = 
		\max \left(
			0, V_{\text{thr}} / \tau_{\text{syn}} - \sum_{k=1}^{K_h} w_{kh} 
		\right)
$}. However, this implicitly assumes that every presynaptic 
weight~$w_{kh}$ is associated with an event~$t_{kh}$. We have previously 
shown~\cite{Pabian2022} that this need not be the case as SNN~can exhibit a sparse 
neuron activity. It is entirely possible that for some neuron $h$ the inequality \mbox{$
	\sum_{k=1}^{K_h} w_{kh} > V_{\text{thr}} / \tau_{\text{syn}}
$} holds and yet the neurons does not fire anyway. This scenario might occur if none of 
the presynaptic neurons observe an event. 
Therefore,~\eqref{eqn:02_methods/loss/spike_penalty_virtual} can be reformulated
to make this dependence on input spikes explicit
\begin{equation}
	R_{\text{spiking}} = \sum_{h=1}^{H} \sum_{m=1}^{M_h} R_{mh}
	\label{eqn:02_methods/loss/spike_penalty_multi} \,,
\end{equation}
where \mbox{$
	R_{mh} = 
		\max \left(
			0, V_{\text{thr}} / \tau_{\text{syn}} - \sum_{k \in B_{mh}} w_{kh} 
		\right)
$} with \mbox{$B_{mh} \subset Q_{mh}$} being the set of valid inputs for the $m$-th output 
of the $n$-th postsynaptic neuron
\begin{equation}
	B_{mh} = 
	\begin{cases}
		\lbrace k: t_{kh} < \infty \rbrace
			& \text{if $m=1$} \\
		\lbrace 
		k: t_{\text{out}_{h}}^{[m-1]} + \tau_{\text{ref}} < t_{kh} < \infty 
		\rbrace
			& \text{otherwise}
	\end{cases}
	\label{eqn:02_methods/loss/valid_inputs_set} \,.
\end{equation}
For completeness
\begin{equation*}
	R_{mh} = 0 \quad \text{if} \quad \lbrace k: t_{kh} < \infty \rbrace = \varnothing
	\,.
\end{equation*}
In our preliminary experiments we found that the penalty term is too strong 
for~\mbox{$m > 1$}, skewing the training objective towards forcing neurons to output 
multiple spikes, rather than letting it focus on solving the actual task. As such, we 
train our models by setting $\forall_{m>1} \, R_{hm} = 0$, which only penalizes neurons 
that do not spike at all (alternatively one might consider applying a smaller 
weight to subsequent spikes).


On another note, the modified cross-entropy loss as defined 
by~\eqref{eqn:02_methods/loss/cross_entropy} ignores the fact that each output layer
neuron can produce multiple spikes. This means that for the last layer, either 
the refractory period could be set to \mbox{$\tau_{\text{ref}}=\infty$}, or that all
$m>1$ output spikes could simply be ignored (as they have no 
impact on the gradient propagation anyway). Both approaches lead to the same result.


Applying the dynamic scaling factor~\eqref{eqn:02_methods/loss/dynamic_penalty} to the
spike regularization term computed over the set of valid 
inputs~\eqref{eqn:02_methods/loss/spike_penalty_multi}, and plugging the result into
the loss function minimized by the original model~\eqref{eqn:02_methods/loss/loss}
leads to the following loss function minimized by the MIMO~SNN training objective
\begin{equation}
	\begin{aligned}
		L_{\text{total}}&(z,y) =
		\frac{1}{N} \sum_{n=1}^{N} L_n(z,y) \\
		&+
		\gamma \frac{1}{\sum_{n=1}^{N} \indicator{y_n \neq \hat{y}_n}}
		\sum_{n=1}^{N} \indicator{y_n \neq \hat{y}_n} R_{\text{spiking}}^{[n]}
		\,.
	\end{aligned}
	\label{eqn:02_methods/loss/mimo_loss} 
\end{equation}

\section{Experimental setup}
\label{sec:03_experiment}

\subsection{Dataset description}
\label{subsec:03_experiment/dataset}

The $\text{RT}_{\text{BUST}}$ study reported in~\cite{mazza2019rtbust} used Twitter 
Premium Search API to compile a list of all Italian retweets shared between 18~June~2018 
and 1~July~2018. In this 2-week period there have been almost 10M retweets shared by 
1.4M distinct users. The compiled dataset\footnote{
	The dataset is available at https://doi.org/10.5281/zenodo.2653137
} consists of records of retweet timestamps
associated with some original tweets. These records can be aggregated based on user id
such that each user is characterized by a different number of tweet-retweet pairs.
Importantly, the dataset contains no information on the content of shared tweets, thus the 
goal is to classify legitimate and bot users based solely on retweet timestamps. 
To supplement this vast collection of unlabeled records, about 1000~accounts were manually 
annotated based on published annotation guidelines for datasets containing social 
bots~\cite{cresci2017paradigm}.

We limit the scope of our analysis to the labeled portion of this dataset and, in 
contrast to the original study, apply a supervised learning algorithm on individual data 
points. Let us introduce a \emph{retweet delay} as a difference between retweet
and the origin-tweet timestamps, expressed in minutes. This quantity can be treated as
describing an observed event, hence we denote it as $t$ in order to avoid introducing
additional notation. We filter the records so that 
only those with both tweet and retweet timestamps occurring from 18~June~2018 to
1~July~2018 (exclusive) are kept, meaning that the retweet delay attribute is bounded 
from above to about $\tau_{\text{max}} = 2 \cdot 10^4$~minutes (2~weeks)~\footnote{
	Throughout the rest of this paper, we express all units of time in terms of minutes, same 
	as the original $\text{RT}_{\text{BUST}}$ study.
}. Additionally, only records 
with a delay of at least $\tau_{\text{min}} = 10^{-1}$ minutes were considered for further 
analysis. Records with a retweet delay below the selected $\tau_{\text{min}}$ can be
considered a sign of either auto-retweeting bots, or legitimate users that were 
refreshing their feed as the tweet was posted and decided to retweet immediately. Either 
way, in our opinion, this does constitute a typical user behavior. As a result of this 
filtering step, 11.73\%~of all records were removed.

Finally, the records were grouped by the retweeting user and by label. This allowed us 
to construct a set of data points $\boldsymbol{X}$ with elements \mbox{
	$\boldsymbol{x} = \left[ t_1, \ldots, t_{U} ; U \right]$
} such that \mbox{$\tau_{\text{min}} < t_1 < \ldots < t_{U} < \tau_{\text{max}}$} 
are the retweet delays bounded by observation time~$\tau_{\text{max}}$ and 
$U=U(\tau_{\text{max}})$~is the length of the sequence. We obtain a 
nearly balanced data subset with 366~examples of legitimate users and 389~examples of 
bot users. These user tweet-retweet sequences have, on average, about 113~events with a 
maximum of~698. The remaining 46,883~cases are unlabeled and were not used in this study.

\subsection{Signal preprocessing}
\label{subsec:03_experiment/preprocessing}

\begin{figure}[t]
	\includegraphics[width=\figurecolumn]{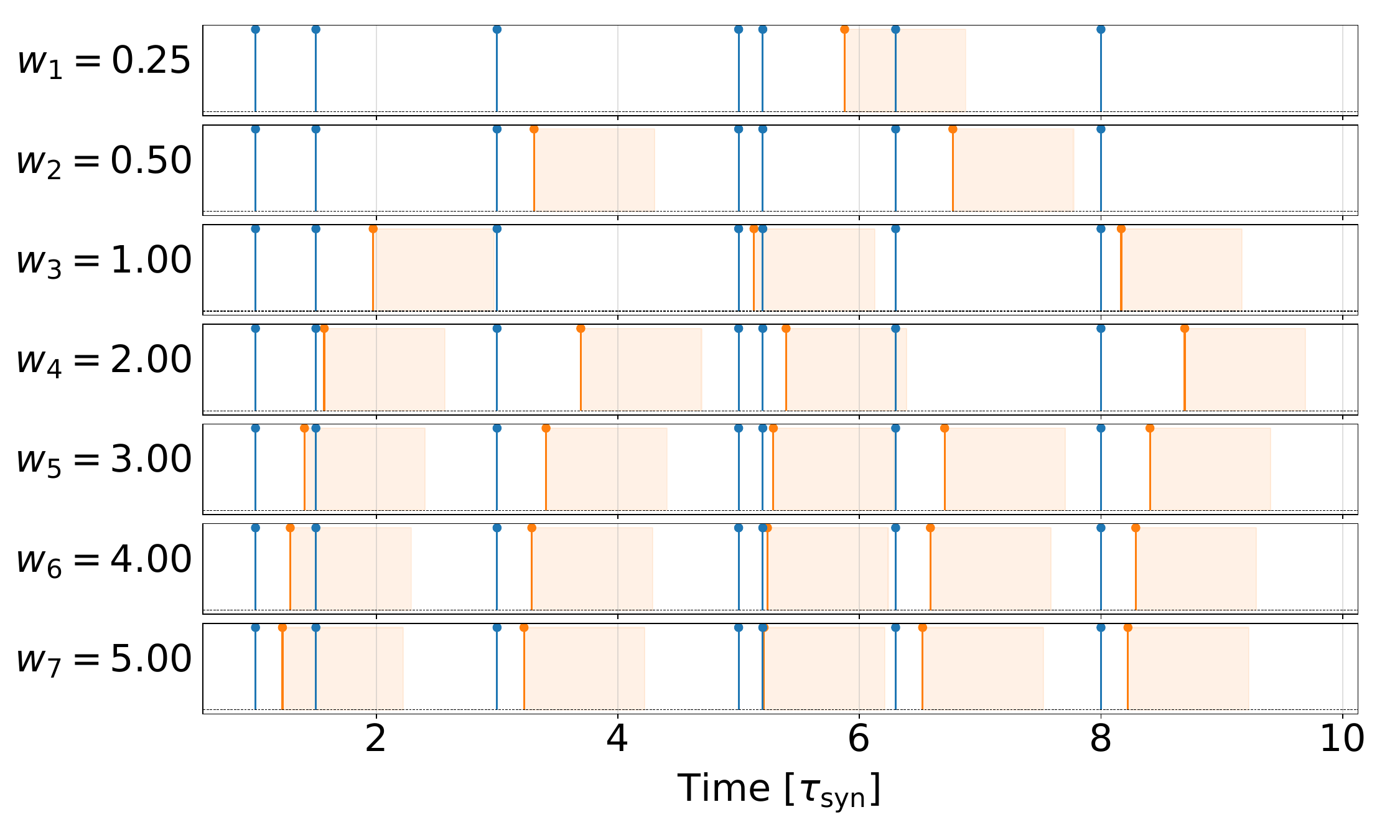}
	\centering
	\caption{
		Simulated spike trains from a simple network with one input neuron and 
		seven postsynaptic neurons. In blue: input spike train (the same in all rows), in 
		orange: spikes generated by postsynaptic neurons. The shaded area denotes the 
		refractory period after generating a spike. All output neurons have the same values 
		of parameters $\tau_{\text{syn}}$, $\tau_{\text{ref}}$ and $V_{\text{thr}}$ with the 
		only difference between them being the synaptic weight $w$. Note that if the weight 
		it too large (in this case $w\ge3$), the neuron elicits a spike in response to every 
		input event, unless it occurs during the refractory period, effectively repeating the 
		input sequence. This means that a group of postsynaptic neurons is redundant because 
		they produce almost identical spike trains. In such scenario, output sequence 
		variability could be improved by adjusting the values of $\tau_{\text{syn}}$, 
		$\tau_{\text{ref}}$ and $V_{\text{thr}}$ for each neuron individually.
	}
	\label{fig:03_experiment/preprocessing/single}
\end{figure}

The tweet-retweet sequences obtained in the previous Section are characterized by two 
peculiar traits that make designing an SNN-based classifier difficult. First, we note 
that all events occur in only one channel. If all neurons in the network have the same value
of parameters $\tau_{\text{syn}}$, $\tau_{\text{ref}}$ and $V_{\text{thr}}$, then this 
scenario imposes additional constraints on the weights during training. First of all, the 
synaptic weight of each connection must be positive, otherwise the postsynaptic neuron is 
unable to produce any spike. This also means that each postsynaptic neuron eventually 
produces a spike as IF~neurons are unable to lose charge if all presynaptic weights are 
positive. Additionally, if some weights are too large, then it is possible that the 
associated postsynaptic neurons will produce nearly identical spike trains, with only slight 
variations in spike frequency and their timing (illustrated in 
\figurename~\ref{fig:03_experiment/preprocessing/single}). While it is certainly not 
impossible to train a network with such constraints on the input layer, we can expect this to 
be more difficult. In fact, in Section~\ref{subsec:04_results/ablation} we empirically show 
that training a model with a single input neuron is indeed more challenging than the 
alternative.

\begin{figure}
	\includegraphics[width=\figurecolumn]{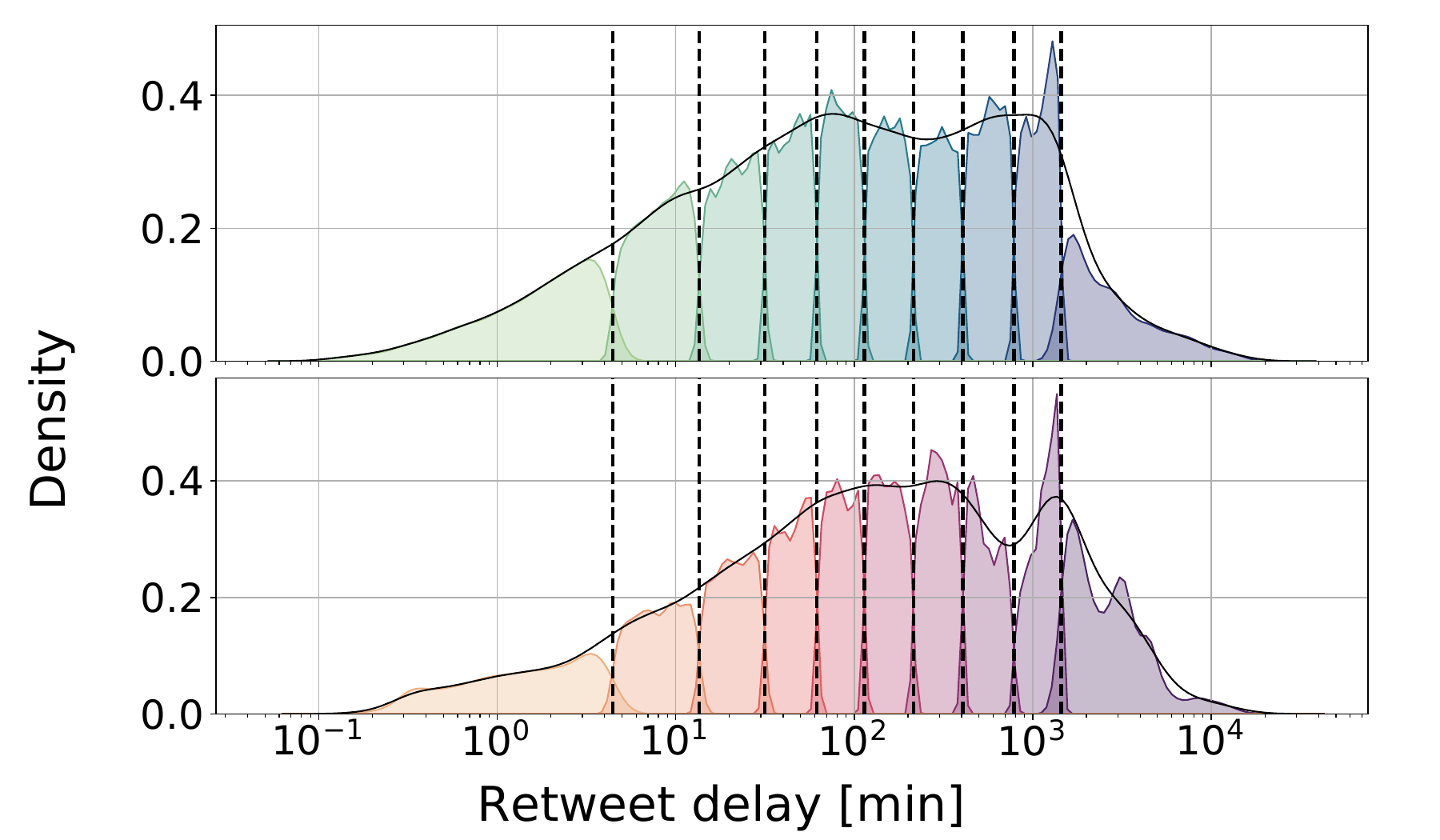}
	\centering
	\caption{The result of binning the empirical density functions of the two classes 
		(above - legitimate users; below - bots) into 10~bins over a given data split.}
	\label{fig:03_experiment/preprocessing/binning}
\end{figure}

For these reasons we explored the possibility of transforming input spike trains into 
multiple sequences of fewer events. This is analogous to conducting feature engineering 
instead of relying on trainable feature extractors in artificial neural networks. 
Loosely inspired by the technique of binning used 
in neuroscientific studies, we identify a collection of 
\emph{bins} that divide the spike train aggregated over all examples into 
sub-sequences with approximately the same number of events in each bin\footnote{
	By contrast, in neuroscience binning produces the number (or an average number of) 
	events that had occurred in a given time interval over multiple repetitions of the 
	experiment.
}. This concept is illustrated in 
\figurename~\ref{fig:03_experiment/preprocessing/binning} by dividing the empirical
density functions of the two classes. Importantly, the step that identifies 
binning thresholds is computed over the training examples. Each sub-sequence can then 
be shifted so that it starts at~$t=0$ by subtracting the corresponding binning 
threshold.

We note that binning is a valid strategy for this specific problem because it is 
assumed that events aggregated into a tweet-retweet sequence of each user are 
independent of one another~\cite{mazza2019rtbust}. While this is a reasonable 
assumption given the data, in reality it seems plausible that graph community 
structure and programmed bot behavior play an important role in what gets retweeted, 
and when. Additionally, the IF~neuron exhibits the memoryless property where the 
internal state of the neuron (membrane voltage) is preserved regardless of how recent 
was the previous event. Finally, as a side note, taking the binning approach to the 
extreme would result in a multitude of thin bins such that there is at most one event per 
bin for each example. We did not consider this to be worth pursuing as it would greatly 
increase the number of neurons needed to represent the signal in the input layer of 
the network.


The second point that needs to be addressed is that events occur at timescales 
differing by almost five orders of magnitude. At such scales the transformed
events in~\eqref{eqn:02_methods/backprop/z_transform} easily surpass the
limits of double-precision floating-point data format. 
This problem still persists even after binning the 
sequences and shifting each sub-sequence so that it has a common starting time 
of~$t=0$. It is easy to see
(\figurename~\ref{fig:03_experiment/preprocessing/binning}) that for bins with a
higher index, the range of event times, while reduced, is still relatively large for 
the network. And so, we can either increase the number of bins (but as discussed 
previously this does not seem to be a suitable approach), or transform each sequence 
to reduce the range of observed values. Therefore, 
we apply a log-transform to binned sub-sequences with events occurring in range \mbox{
	$t \in \left[T_{c-1}, T_{c}\right)$
} such that
\begin{equation}
	\forall_{t \in \left[T_{c-1}, T_{c}\right)} \enspace 
		g(t; b_c, c) = \log_{b_{c}} \left( t - T_{c-1} + 1 \right)
	\label{eqn:03_experiment/preprocessing/logscaling}
\end{equation}
for $b_c > 1$, where~$c \in \lbrace 1, 2, \ldots \rbrace$ is the index of the bin, 
$T_{c}$ is the upper boundary of the~$c$-th bin (i.e., the threshold), and $T_0 = 0$. 
Each bin corresponds to a single network input channel, hence
we reuse the index $c$ to make this explicit and avoid introducing additional notation.
Note that each transformed sequence is shifted to start 
at \mbox{$t=0$}. This transform is controlled by a single parameter~$b_c$, the base of 
the logarithm. Note that $g(t; b_c, c)$ is a decreasing function of $b_c$.

In general, the transform in~\eqref{eqn:03_experiment/preprocessing/logscaling} allows 
setting a different~$b_c$ for each bin (channel). In our experimental scope we consider 
two strategies for selecting~$b_c$:
\begin{enumerate}
	\item Setting the same base $b > 1$ for all bins
	\begin{equation}
		\forall_{t \in \left[T_{c-1}, T_{c}\right)} \enspace 
			g_{1}(t; b, c) = \log_{b} \left( t - T_{c-1} + 1 \right) \,.
		\label{eqn:03_experiment/preprocessing/logscaling_1} 
	\end{equation}	
	
	\item Adjusting the base of the logarithm for each bin separately so that all bins have 
	the same range of values after transform
	\begin{equation}
		g_{2}(t; \kappa, c) = \frac{\kappa}{g_{1}(T_c; b, c)} g_{1}(t; b, c)
		\label{eqn:03_experiment/preprocessing/logscaling_2} \,,
	\end{equation}	
	where~$\kappa$ is the time-instant assigned to the threshold~$T_{c}$ after 
	transform, i.e., \mbox{$\kappa = g_{2}(T_{c}; \kappa, c)$}. Note that in this strategy 
	the value of the parameter $b$ does not matter as the actual logarithm base for a given 
	bin~$c$ is controlled by the boundaries $T_{c-1}$, $T_{c}$ and the 
	parameter~$\kappa$. For simplicity, we use the same~$\kappa$ for all bins.
\end{enumerate}
The difference between these two approaches lies in the range of values produced by the 
transform. Strategy $g_{1}(t; b, c)$ is unbounded from above and so it might be difficult to 
select a single~$b$ that works at both short and long timescales. Conversely, the function
$g_{2}(t; \kappa, c)$ squeezes all values to lie in the range $\left[0, \kappa\right]$, which 
might make it easier for the network to learn the relationship between events at different 
timescales.

\begin{figure}
	\includegraphics[width=\figurecolumn]{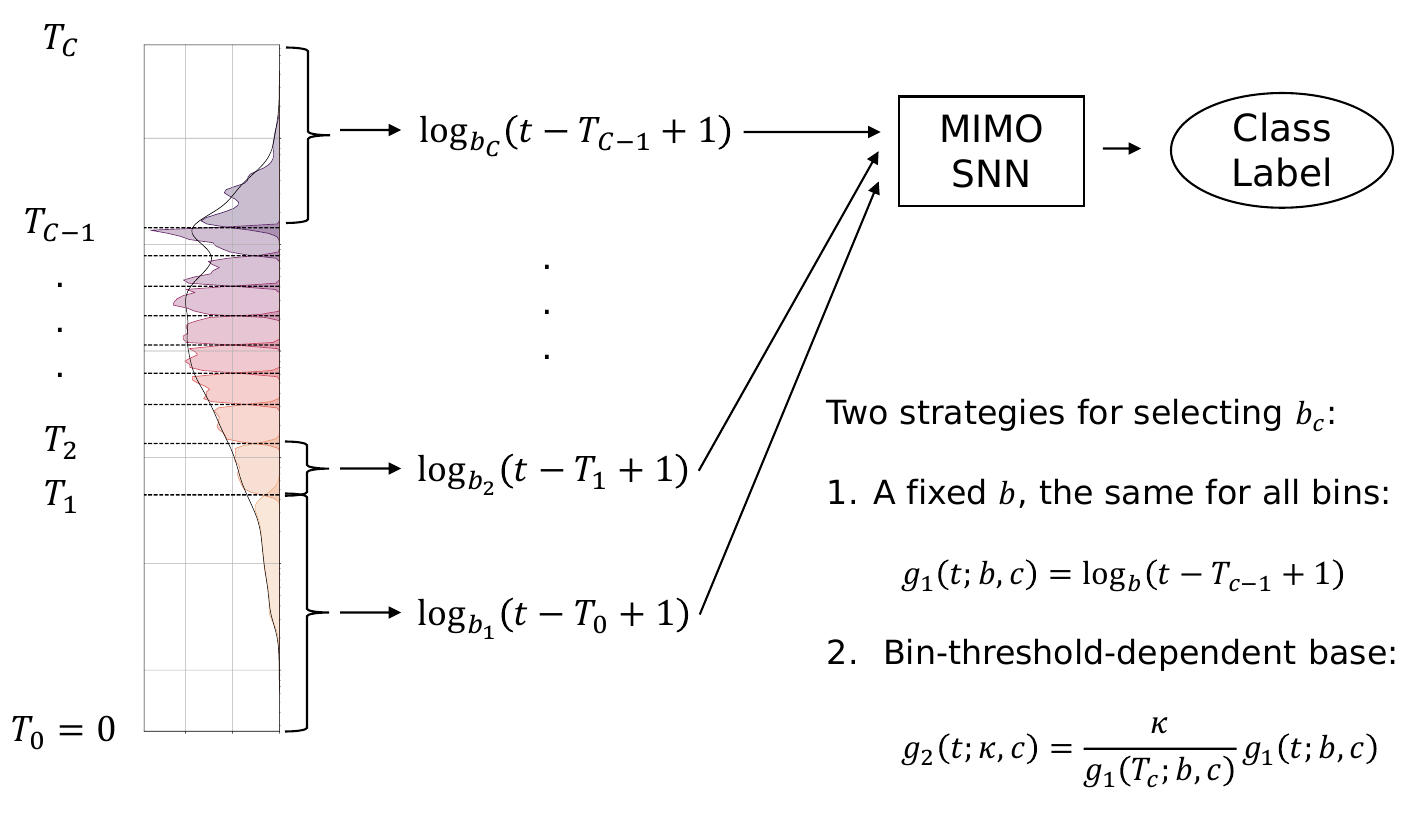}
	\centering
	\caption{A diagram of the preprocessing steps to obtain signals used to train a 
		MIMO~SNN on Twitter dataset.}
	\label{fig:03_experiment/preprocessing/combined}
\end{figure}

\figurename~\ref{fig:03_experiment/preprocessing/combined} summarizes the outlined
preprocessing steps. We found that applying the~$b$-parameterized log-transform on 
binned sub-sequences sufficiently addressed our concerns for training the proposed 
spiking neural network. Note that while the 
log-transform~\eqref{eqn:03_experiment/preprocessing/logscaling}
has the unfortunate effect of significantly increasing the latency of the model, we are only 
interested in the final classification prediction of the model and not its real-time 
performance.

\section{Results \& discussion}
\label{sec:04_results}

In all of our experiments we used the same \mbox{$C$-12-24-48-2} architecture, 
where~$C$ is the number of bins used to preprocess the spike train, the network has 2~output 
neurons, and other digits represent the number of neurons in the hidden layers. This 
produces a relatively small network of about $1536+12C$ parameters. However, a MIMO~SNN 
model complexity is not only determined by the number of parameters, but also by the 
refractory period~$\tau_{\text{ref}}$ (see~\eqref{eqn:02_methods/mimo/causal_set_modified}) 
which controls how often each neuron in the network is able to spike.

\subsection{The impact of the refractory period on signal propagation}
\label{subsec:04_results/preliminary}

In order to settle on a single value of~$\tau_{\text{ref}}$
to use in our experiments, we conducted a preliminary study by training the 
classification model on a small subset of data (stratified random $10\%$~sample). 
We considered \mbox{$\tau_{\text{ref}} \in \left[0.01, 1\right]$}, selected on a 
5-point logarithmically-spaced grid. The data was preprocessed into 30~bins, with 
bin-threshold-dependent logarithm base~$b_c$ computed for $\kappa=3$. All models were 
trained for 200~epochs of 10~update steps each, with a batch size of~64. This was enough
to reach perfect classification score on the training set for each model
(note that in this experiment we were not interested in the classifier's performance on
unseen data, but rather in how the signal propagates through the MIMO network).
The synaptic regularization parameter was set to $\gamma=10^5$ in order to promote spiking 
activity in the network. Training was repeated 5~times in order to average-out processing 
time measurements.

In the preliminary study we measured the average time it took to finish the training 
epoch, as well as sparsity-adjusted average number of spikes produced by the network in 
response to input signals. The ``sparsity-adjusted'' term states that only the 
active (i.e., not quiescent) neurons are considered when discussing the impact of the 
choice of~$\tau_{\text{ref}}$. To succinctly describe this measure we introduce
the \emph{network activity indicator}
\begin{equation}
	\mathrm{NAI} = 
		\frac{1}{N \sum_{l=1}^{L} H_l} 
		\sum_{n=1}^{N} \sum_{\ell=1}^{L} \sum_{h_{\ell}=1}^{H_{\ell}} M_{nh_{\ell}}
	\label{eqn:04_results/preliminary/nai} \,,
\end{equation}
where $M_{nh_{\ell}}$ is the number of output spikes generated by the $h_{\ell}$-th neuron 
of the $\ell$-th layer in response to the $n$-th example in the minibatch.

\begin{figure}[t]
	\includegraphics[width=\figurecolumn]{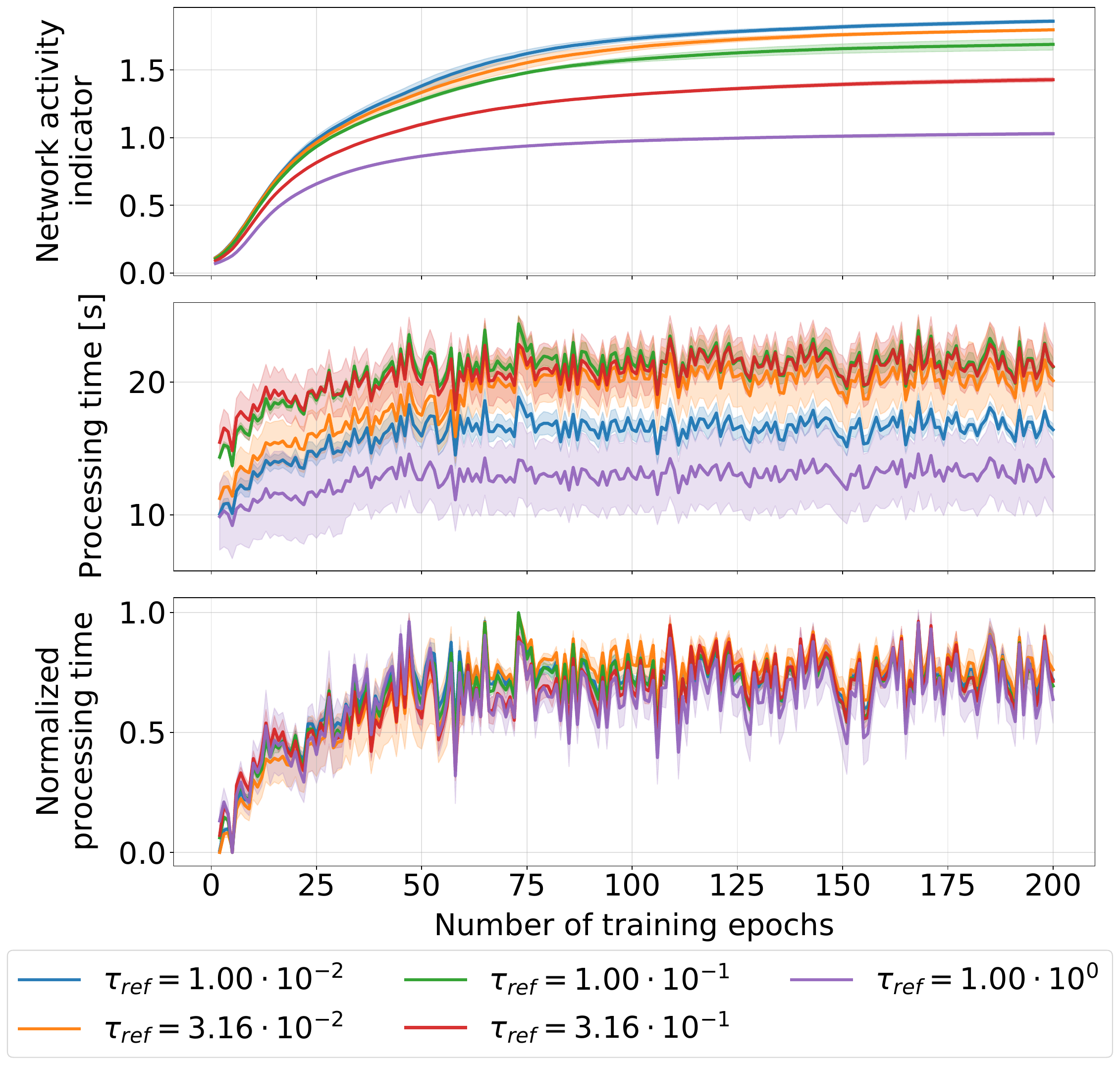}
	\centering
	\caption{
		The impact of the refractory period~$\tau_{\text{ref}}$ on a trained 
		MIMO~SNN properties. 
		Top panel:~sparsity-adjusted average number of spikes produced by the network. 
		Middle panel:~average epoch training time. 
		Bottom panel:~normalized average epoch training time.
	}
	\label{fig:04_results/preliminary/curves}
\end{figure}

The results are summarized in~\figurename~\ref{fig:04_results/preliminary/curves}. 
We observe that both the NAI~measure and the epoch training time rapidly increase in the 
initial stages (the first 50~epochs) because then the spike regularization term forces the 
model to learn how to propagate the signal through the network. Afterwards, the training 
time remains relatively constant throughout the rest of the training, while NAI~continues 
to increase, albeit at a much slower rate. Note that the growth of~NAI over training epochs 
is unbounded because there is no term in the loss 
function~\eqref{eqn:02_methods/loss/mimo_loss} that encourages the network to generate 
fewer spikes. This shows that setting a smaller~$\tau_{\text{ref}}$ causes the 
network to produce more spikes. 

Surprisingly, however, the obtained processing time 
measurements do not support the notion that it is possible to predict which model will take 
the least amount of time to process the examples based solely on the value of the 
refractory period. Such relationship was anticipated as the number of iterations needed to 
compute all output spikes in~\eqref{eqn:02_methods/mimo/if_zout_modified} increases as 
$\tau_{\text{ref}}$~decreases. Perhaps repeating this experiment significantly more times 
would allow us to reach a conclusive answer as processing time measurements are notoriously 
unreliable. Nevertheless, normalizing the processing time measurements separately for each 
experimental run shows that the relative increase in processing time over the training 
epochs is similar across different scenarios. Based on all these results, we settle 
on~$\tau_{\text{ref}}=0.1$ in our further experiments as the largest~$\tau_{\text{ref}}$ 
that still exhibits a steep increase in NAI~measure over training steps.

\subsection{Bot detection - binary classifier performance}
\label{subsec:04_results/classification}


In our experiments on the classification problem, having fixed the value 
of~$\tau_{\text{ref}}$, we explored the impact of proposed preprocessing on model 
performance. We constructed a grid over preprocessing parameter space, selecting the 
number of bins from the set \mbox{$C \in \lbrace 10, 20, 30, 40, 50 \rbrace$}, whereas the 
log-transform parameter was set to either~$b_{c}(\kappa)$ for \mbox{
	$\kappa \in \lbrace 1, 2, 3 \rbrace$
}, or \mbox{$b \in \lbrace 10, 30 \rbrace$}. For every pair of parameters on this 
grid, several models were trained and evaluated with~$5$-fold cross validation. Thus, 
in total $125$~models were trained. All models were trained for $50$~epochs, each with 
$100$~training steps over $64$~class-balanced training examples. The learning rate was
set to~$10^{-3}$ in all steps. The synaptic regularization term
factor~$\gamma$ varied substantially depending on the current training epoch -- 
initially set to~$10^{5}$ in order to guide the model towards a state in which it is 
able to propagate spikes throughout all layers. After $10$~epochs, we 
set~$\gamma=10^{-2}$ so that the model could focus on minimizing the task-specific loss 
term~\eqref{eqn:02_methods/loss/cross_entropy}.


Given the small dataset size (366~legitimate users and 389~bots), we opt to use data 
augmentation in order to increase the effective training split size. Two types of 
augmentation were used:
\begin{itemize}
	\item drop events with probability~$0.1$,
	\item randomly shift each event in time by $t_{\delta}$ uniformly distributed in 
	$(-0.05, 0.05)$, independently with probability~$0.3$.
\end{itemize}
The latter augmentation type is applied only after the sequence has been preprocessed, 
making sure that the shift-augmented sub-sequence is still composed of events 
occurring at nonnegative time.


The classification accuracy achieved by all models trained over the preprocessing grid 
is summarized in~\tablename~\ref{tab:04_results/classification/grid_eval}. The results
obtained during the hyperparameter search on the preprocessing grid do not suggest an
existence of some pattern that holds across different number of bins or the transform 
choice. Notably, however, the setting denoted by the parameter~$b=30$ seems to be more 
robust compared to other settings in that it is the only one that allowed the model to 
consistently reach an accuracy over~70\% regardless of the number of bins. 

Overall,
the wide range of performance classification scores reported by different models
suggest that it is imperative to perform hyperparameter tuning when using the proposed
MIMO~network for this specific scenario of extremely long-range dependencies between 
events. Nevertheless, it must be stressed that the hyperparameters associated with the 
neural model itself ($V_{\text{thr}}$, $\tau_{\text{syn}}$, $\tau_{\text{ref}}$) can be 
selected heuristically. Firstly, note that $V_{\text{thr}}$ only influences the scale 
of trained weights and has no impact on training dynamics, making the choice of 
$V_{\text{thr}}$ arbitrary. On the other hand, the time constants $\tau_{\text{syn}}$ 
and $\tau_{\text{ref}}$ can be selected according to the input event distribution. The 
post-synaptic potential constant $\tau_{\text{syn}}$ needs to be longer than the 
relevant temporal patterns present in the input data~\cite{bohte2002error}. Lastly, 
setting $\tau_{\text{ref}} < \tau_{\text{syn}}$ prevents the scenario in which the 
network rarely generates events.

\tablename~\ref{tab:04_results/classification/comparison} compares our best model with the 
results obtained in the original study~\cite{mazza2019rtbust} in terms of accuracy, 
recall, precision, $F_1$-score and Matthews correlation coefficient~(MCC). Our MIMO~SNN 
model outperforms all of the presented supervised approaches (Botometer, Social 
fingerprinting), as well as most of the unsupervised methods (HoloScope; 
PCA- and TICA-based $\mathrm{RT_{BUST}}$). Importantly, the latter group of methods relied 
on fitting the model on the entire unlabeled dataset of 63,762~accounts and evaluating on 
the labeled portion, whereas we focus only on the annotated subset (as outlined in
Section~\ref{subsec:03_experiment/dataset}), composed of about 755 labeled cases in total.
We note that the Variational Autoencoder~(VAE) variant of the $\mathrm{RT_{BUST}}$ 
model performed better than our approach. As this 
model also leveraged the unlabeled portion of the dataset, it stands to reason that there 
is a clear benefit to clustering-based methods, in which the presence of a suspicious 
behavior emerges only when analyzing user in groups, rather then as individuals. 
Furthermore, we note that the examples examined in~\cite{mazza2019rtbust} were manually 
labeled and thus any incorrectly labeled cases could have had a much bigger impact on the
supervised model trained with significantly fewer examples.
Lastly, it is important to note that the original study lacks technical details 
related to the LSTM-VAE network architecture, which prevents us from making a comparison 
between the two methods that is adjusted for model complexity as expressed by the number 
of trainable parameters.

\begin{table}
	\hfill
	\caption{Classification accuracy for MIMO SNN models trained on Twitter 
		tweet-retweet dataset, depending on the chosen preprocessing parameters.
		A star ($\star$) denotes the best result for a row, whereas the diamond 
		($\diamond$) denotes the best result in a column.
	}
	\label{tab:04_results/classification/grid_eval}
	\begin{center}
		\tiny
		\renewcommand{\arraystretch}{1.3}
		\begin{tabular}{cc|ccccc|}
			\cline{3-7}
			\multicolumn{2}{c|}{\multirow{2}{*}{\makecell{Stratified 5-fold\\cross validation}}} & \multicolumn{5}{c|}{\thead{Log-transform-related parameter}}                                                                                                                                 \\ \cline{3-7} 
			\multicolumn{2}{c|}{}                                                    & \multicolumn{1}{r|}{$b=10$}  & \multicolumn{1}{r|}{$b=30$}  & \multicolumn{1}{r|}{$b_{c}(\kappa=1)$}  & \multicolumn{1}{r|}{$b_{c}(\kappa=2)$}           & \multicolumn{1}{r|}{$b_{c}(\kappa=3)$}           \\ \hline
			\multicolumn{1}{|c|}{\multirow{5}{*}{\rotatebox[origin=c]{90}{\thead{Number of bins}}}}      & $C=10$      & \multicolumn{1}{r|}{$^{\star\diamond}73.25 \pm 3.71$} & \multicolumn{1}{r|}{$70.33 \pm 3.83$} & \multicolumn{1}{r|}{$68.74 \pm 3.09$} & \multicolumn{1}{r|}{$68.08 \pm 4.20$} & \multicolumn{1}{r|}{$68.61 \pm 10.46$} \\ \cline{2-7} 
			\multicolumn{1}{|c|}{}                                     & $C=20$      & \multicolumn{1}{r|}{$70.20 \pm 3.72$} & \multicolumn{1}{r|}{$^{\star}70.60 \pm 1.65$} & \multicolumn{1}{r|}{$67.95 \pm 2.70$} & \multicolumn{1}{r|}{$70.33 \pm 3.80$} & \multicolumn{1}{r|}{$68.21 \pm \phantom{0}5.70$} \\ \cline{2-7} 
			\multicolumn{1}{|c|}{}                                     & $C=30$      & \multicolumn{1}{r|}{$67.42 \pm 2.03$} & \multicolumn{1}{r|}{$^{\star\diamond}71.52 \pm 4.74$} & \multicolumn{1}{r|}{$69.40 \pm 4.76$} & \multicolumn{1}{r|}{$70.86 \pm 1.83$} & \multicolumn{1}{r|}{$69.14 \pm \phantom{0}3.09$} \\ \cline{2-7} 
			\multicolumn{1}{|c|}{}                                     & $C=40$      & \multicolumn{1}{r|}{$69.27 \pm 2.67$} & \multicolumn{1}{r|}{$70.60 \pm 4.40$} & \multicolumn{1}{r|}{$66.75 \pm 4.50$} & \multicolumn{1}{r|}{$69.40 \pm 4.42$} & \multicolumn{1}{r|}{$^{\star\diamond}71.13 \pm \phantom{0}3.49$} \\ \cline{2-7} 
			\multicolumn{1}{|c|}{}                                     & $C=50$      & \multicolumn{1}{r|}{$67.68 \pm 4.20$} & \multicolumn{1}{r|}{$71.26 \pm 6.22$} & \multicolumn{1}{r|}{$^{\star\diamond}72.19 \pm 3.82$} & \multicolumn{1}{r|}{$^{\diamond}70.99 \pm 4.26$} & \multicolumn{1}{r|}{$68.08 \pm \phantom{0}3.61$} \\ \hline
		\end{tabular}
	\end{center}
	\hfill
\end{table}

\begin{table}
	\hfill
	\caption{
		Comparison of model performance on the bot detection task between different 
		techniques.
	}
	\label{tab:04_results/classification/comparison}
	\begin{center}
		\tiny
		\begin{tabular}{|l|l|r|r|r|r|r|}
			\hline
			\textbf{}      & \textbf{Technique}                                             & \textbf{Accuracy} & \textbf{Recall}  & \textbf{Precision} & $\mathbf{F_1}$\textbf{-score} & \textbf{MCC}    \\ \hline
			\multirow{3}{*}{\cite{mazza2019rtbust}} & Botometer                                   & $58.30$           & $30.98$          & $69.51$            & $42.86$                       & $0.2051$        \\
	                                                & HoloScope                                   & $49.08$           & $0.49$           & $28.57$            & $0.96$                        & $-0.0410$       \\
	                                                & Social fingerprinting                       & $71.14$           & $89.78$          & $65.62$            & $75.82$                       & $0.4536$        \\ \hline
			\multirow{4}{*}{\cite{mazza2019rtbust}} & $\mathrm{RT_{BUST}}$ (PCA)                  & $51.54$           & $95.12$          & $51.11$            & $66.49$                       & $0.0446$        \\
													& $\mathrm{RT_{BUST}}$ (TICA)                 & $53.64$           & $95.12$          & $52.28$            & $67.47$                       & $0.1168$        \\
													& $\mathrm{RT_{BUST}}$ (VAE)                  & $87.55$           & $81.46$          & $93.04$            & $86.87$                       & $0.7572$        \\ \hline
			our                                     & MIMO SNN                                    & $73.25 \pm 3.71$  & $69.56 \pm 8.93$ & $76.39 \pm 3.34$   & $72.46 \pm 5.08$              & $0.47 \pm 0.07$ \\ \hline
		\end{tabular}
	\end{center}
	\hfill
\end{table}

\subsection{Model ablation study}
\label{subsec:04_results/ablation}


For the best-performing model, we ran an ablation study experiment in order to determine
which components of the proposed approach have a significant impact on model performance.
We tested four scenarios:
\begin{enumerate}
	\item no data augmentation,
	\item setting $\tau_{\text{ref}}=\infty$, i.e., preventing the neurons in the network
	from spiking more than once in response to a given input signal,
	\item no splitting of channels into separate bins,
	\item no log transform.
\end{enumerate}
Each one represents a single change to the experimental protocol outlined in 
Section~\ref{subsec:04_results/classification}. Clearly, with the exception of the first 
scenario, these changes have a major impact on how the signal is propagated through the 
network or on training dynamics.

As evidenced in \tablename~\ref{tab:04_results/ablation}, removing any of the components 
causes a reduction in model performance. This shows that the proposed data 
augmentation scheme was effective in mitigating the problem of overfitting. Furthermore, 
the drop in performance for the $\tau_{\text{ref}}=\infty$ scenario suggests that some of
the model's capacity to process information is tied to the ability to process it over time. 
Unsurprisingly, not splitting the input spike train into multiple bins makes it more 
difficult for the model to learn long range dependencies, which is an effect that was 
anticipated while designing this preprocessing step. Lastly, the sharpest drop in 
performance is observed when the data is passed through the network without any transform 
that squashes the range of values at its input. Note that the obtained average accuracy 
of~47.68\% is quite close to the ratio of the number of legitimate users to all users in 
the stratified evaluation data split~(48.48\%). This means that the model was unable to 
learn anything, most likely because using raw event times as large as~$2 \cdot 10^4$ minutes
transformed according to~\eqref{eqn:02_methods/backprop/z_transform} surpasses the 
limits of double-precision floating-point data format, making training impossible.
Note that the trained models are in general biased towards higher precision, with the 
exception of the channel-splitting scenario, despite being trained with a nearly 
class-balanced data. In bot-detection scenarios it is preferable to favor precision 
instead of recall as the system administrator should be reasonably certain that a user
is a bot before taking action.

\begin{figure}[t]
	\captionsetup{position=top}
	\subfloat[baseline]{\includegraphics[scale=\figureconfusion]
		{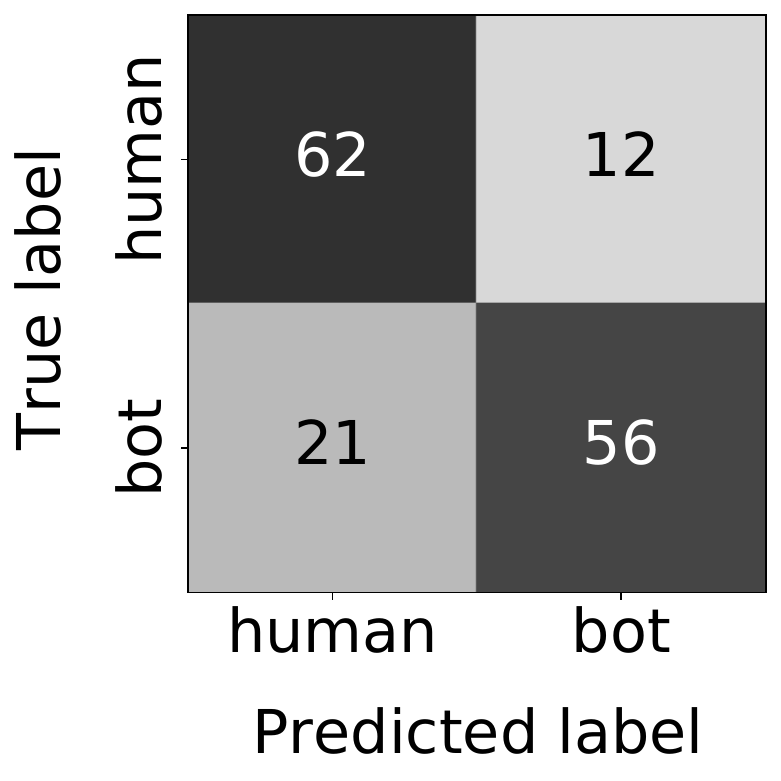}
	}\quad\quad\quad\quad
	\subfloat[no data augm.]{\includegraphics[scale=\figureconfusion]
		{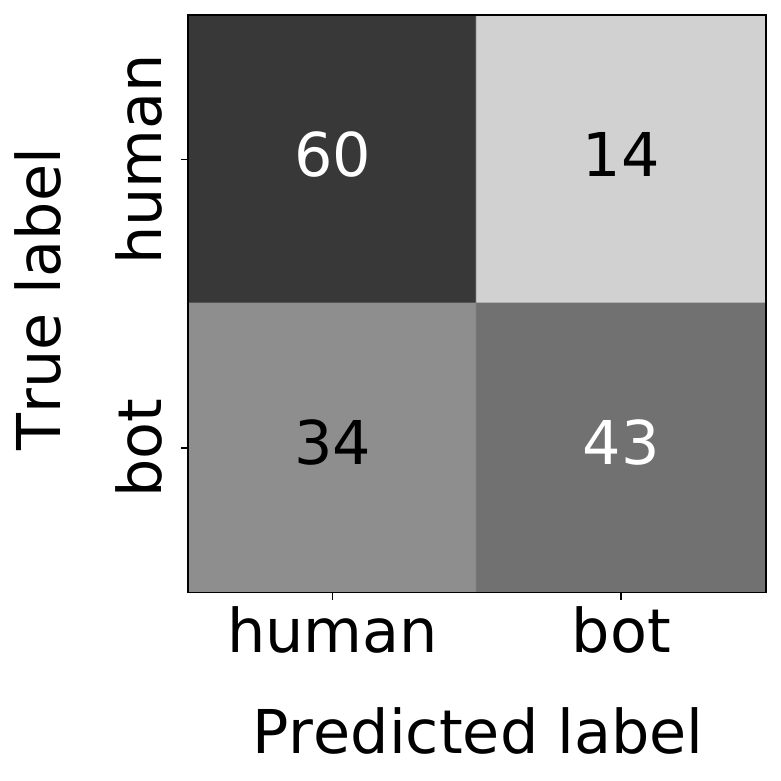}
	}\quad\quad\quad\quad
	\subfloat[inf. refr. period]{\includegraphics[scale=\figureconfusion]
		{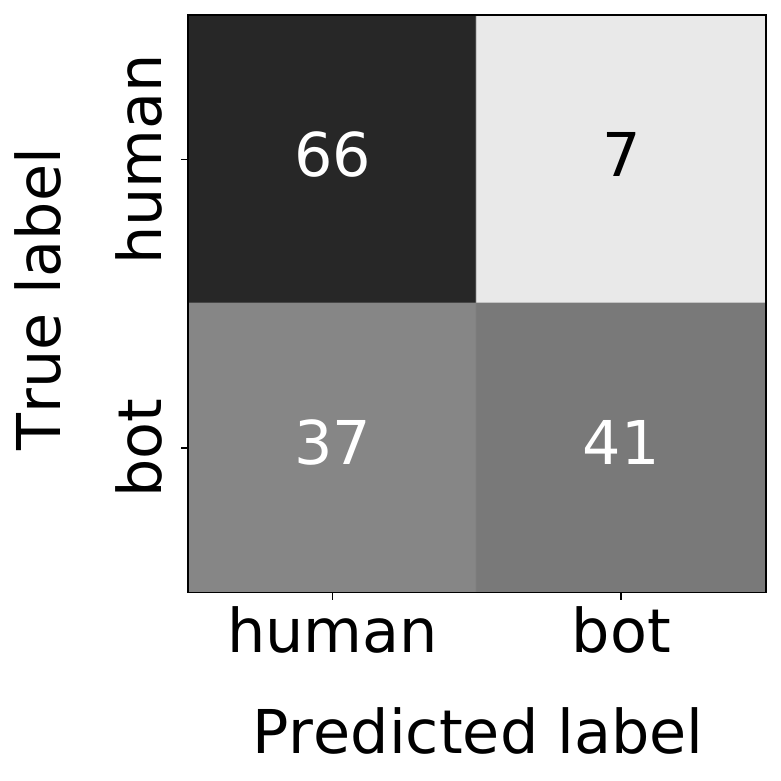}
	}\quad\quad\quad\quad
	\subfloat[no channel-split.]{\includegraphics[scale=\figureconfusion]
		{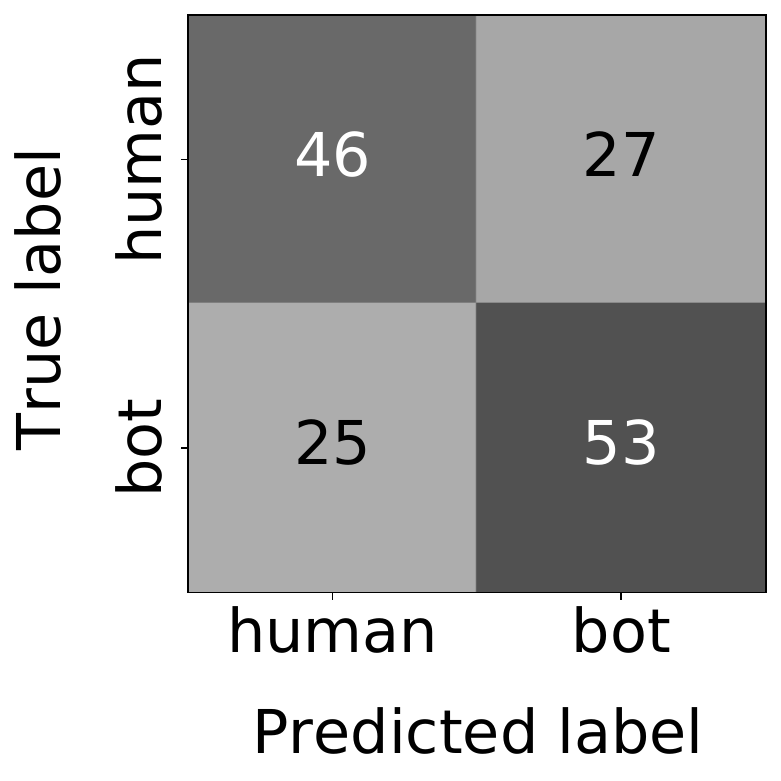}
	}\quad\quad\quad\quad
	\subfloat[no log transform]{\includegraphics[scale=\figureconfusion]
		{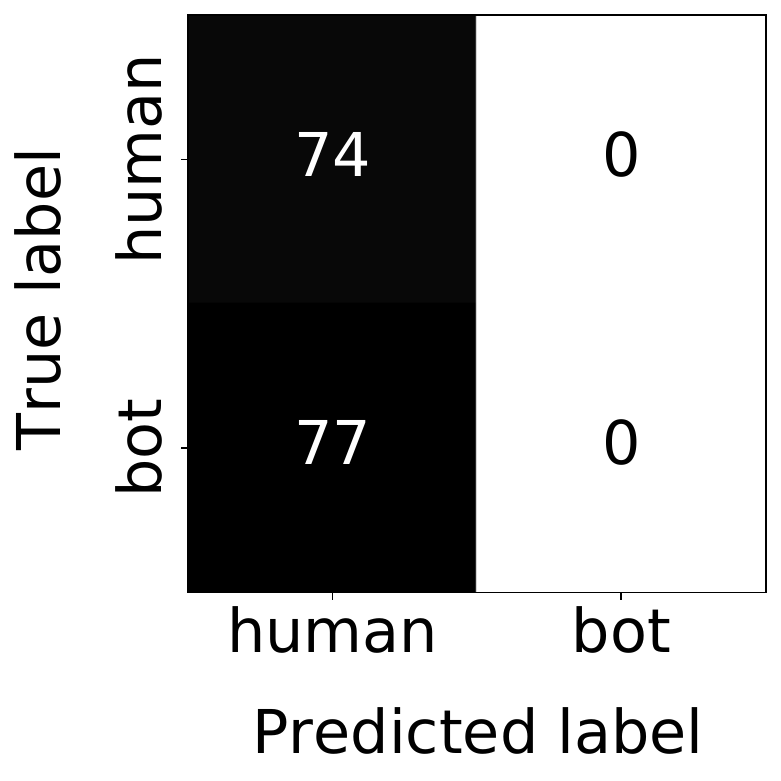}
	}
	\centering
	\caption{
		Confusion matrices corresponding to the best-performing iteration with respect to 
		MCC value for five ablation experiment scenarios.
	}
	\label{fig:04_results/ablation/confusion}
\end{figure}

For completeness, \figurename~\ref{fig:04_results/ablation/confusion} shows representative 
examples of confusion matrices for the five scenarios, chosen according to the~MCC. It is 
evident in the results for the experiment without log transform that for a failed model 
all predictions are towards the negative class. This stems from the assumption that if  
the output layer produces no events, then the model should return the negative class
prediction. However, this effect is negligible for properly trained models -- in other 
scenarios less than 1\% of bot accounts were misclassified as legitimate users due to 
this assumption.

\begin{table}
	\hfill
	\caption{Model performance in the ablation study, given the specified scenario.}
	\label{tab:04_results/ablation}
	\begin{center}
		\tiny
		\begin{tabular}{|l|r|r|r|r|r|}
			\hline
			\multicolumn{1}{|l|}{\textbf{Scenario}} & \multicolumn{1}{r|}{\textbf{Accuracy}} & \multicolumn{1}{r|}{\textbf{Recall}} & \multicolumn{1}{r|}{\textbf{Precision}} & \multicolumn{1}{r|}{$\mathbf{F_1}$\textbf{-score}} & \multicolumn{1}{r|}{\textbf{MCC}} \\ \hline
			baseline                                & $73.25 \pm 3.71$                       & $69.56 \pm \phantom{0}8.93$          & $76.39 \pm 3.34$                        & $72.46 \pm 5.08$                                   & $0.47 \pm 0.07$                   \\ \hline
			no data augmentation                    & $66.89 \pm 1.45$                       & $58.09 \pm 10.61$                    & $73.77 \pm 6.50$                        & $63.80 \pm 5.60$                                   & $0.36 \pm 0.03$                   \\
			infinite refractory period              & $67.28 \pm 2.64$                       & $56.30 \pm \phantom{0}4.52$          & $74.44 \pm 5.63$                        & $63.89 \pm 3.06$                                   & $0.36 \pm 0.06$                   \\
			no channel-splitting                    & $63.18 \pm 4.60$                       & $69.67 \pm 14.22$                    & $62.61 \pm 2.74$                        & $65.42 \pm 7.49$                                   & $0.27 \pm 0.10$                   \\
			no log transform                        & $47.68 \pm 1.18$                       & $0.00 \pm \phantom{0}0.00$           & $0.00 \pm 0.00$                         & $0.00 \pm 0.00$                                    & $-0.06 \pm 0.07$                  \\ \hline
		\end{tabular}
	\end{center}
	\hfill
\end{table}

\subsection{Computational complexity of the MIMO SNN}
\label{subsec:04_results/complexity}

Let us briefly comment on the computational burden of the MIMO~SNN in general, without 
the preprocessing steps that are specific to the Twitter dataset. 
The proposed algorithm slightly increases the complexity of the single-spike
version proposed in~\cite{Mostafa2018}. Note that it is possible to compute the first
output spike of all output neurons in a given layer observing numerous minibatch input 
examples presented to presynaptic neurons in a single pass through the layer 
(see~\cite{zhou2021temporal} for a simplified version of this algorithm). 
While the proposed MIMO~SNN approach introduces an unavoidable feedback loop resulting 
from the spike causality principle, it does not prevent the model from training
with modern deep learning software frameworks. During such iterative search over output 
spikes, the SNN is no different from a nonspiking RNN.

Note that the proposed 
algorithm iterates over the space of discrete output spike events rather than 
discretizing time at a predefined time-resolution. This leads to significantly fewer 
iterations when computing the full output spike train of each neuron than the 
alternative requiring a discretized time simulation. Additionally, this allows MIMO~SNN 
to compute arbitrarily late events w.r.t. reference time, whereas discretized-time 
algorithms are limited in scope by the simulation window. In our experiments we did not 
impose an upper limit on the number of events generated by a single neuron, relying 
instead on the exhaustive search over the event space. Note that this event space is 
finite as each neuron cannot generate more events than it observes across all synapses.

Nevertheless, the MIMO~SNN adversely scales with the number of events observed by 
presynaptic neurons, as well as by those generated by its neurons. The former has the 
biggest effect on the input layer, in which the number of virtual (time-flattened) 
channels can be extremely large when the network has numerous input neurons, each 
observing lengthy spike trains. This limits the applicability of our approach to 
multichannel data streams without too many events (although it is difficult to estimate 
what this upper limit actually is), unless the dataset is preprocessed to contain fewer 
events. Furthermore, the average processing time increases with the number of spikes 
generated by the layer as the feedback loop must be unrolled over time. However, note 
that each successive spike is less likely to be generated (because the causal set 
shrinks with each iteration), making the computational complexity of subsequent 
iterations smaller than the preceding one. Fortunately, the network spiking activity 
can be controlled by the $\tau_{\text{ref}}$ hyperparameter, as evidenced by our 
experiments.

\section{Conclusion}
\label{sec:05_conclusions}

This study introduces a MIMO~SNN model which extends existing time-coding single-spike 
time-to-first-spike SNN models by defining the rules for spike 
propagation throughout the layers when simulating the network on conventional
hardware. Once trained, the model can be realized on existing neuromorphic devices that 
implement the IF neuron computation. In contrast to other works, the proposed MIMO~SNN 
expresses the entire algorithm in terms of iteratively calculating successive spikes,
which stands in contrast to other works simulating the state of the entire network over
a finite time window with a fixed time step. The proposed MIMO~SNN is suited towards
datasets with spike trains composed of relatively infrequent events occurring at different
timescales, such as the Twitter user activity dataset. Furthermore, the algorithm can be
implemented in modern deep learning frameworks, making it scalable to large volumes of 
data.

The model was applied to a labeled subset of Twitter user activity data in order 
to determine whether each analyzed user is legitimate or not. 
Choosing a dataset with a time horizon of a 2-week-period allowed us to show  
that the model is able to process time series composed of events occurring at 
timescales differing by almost five orders of magnitude. Our best model achieved an
accuracy score of~$73.25\%$, compared to~$87.55\%$ obtained by the original 
$\mathrm{RT_{BUST}}$ study. However, we note that the latter is an unsupervised 
learning algorithm and is therefore able to infer different non-overlapping patterns of 
activity of distinct groups of users not present in the class-label-aggregated data. 
Furthermore, our model relies on about 755~labeled example compared to the
63,762~unlabeled cases used to train $\mathrm{RT_{BUST}}$. As the data was manually
labeled by the authors of
the study, any incorrectly labeled examples could have had a much bigger impact on the 
supervised model trained with significantly fewer examples.
We have shown that the proposed MIMO~SNN operates directly in an event-domain, and so there 
is no need to encode the time series in any way for it to be processed by the model. In 
addition to the classification model feasibility study, this work showcased novel signal 
preprocessing steps, exemplary spike train data augmentation techniques, and the 
heuristic of modifying regularization scale factor during training to tackle this 
challenging dataset. We found that these 
concepts were critical at preventing overfitting and stabilizing the training procedure,
as evidenced by the results of the ablation study. We hope to explore these insights in 
future works.

\bibliographystyle{elsarticle-num}
\bibliography{references}

\end{document}